\documentclass{article} 
\usepackage{iclr2027_conference,times}

\usepackage{amsmath,amsfonts,bm}

\def\eqref#1{equation~\ref{#1}}

\def\1{\bm{1}}

\DeclareMathAlphabet{\mathsfit}{\encodingdefault}{\sfdefault}{m}{sl}
\SetMathAlphabet{\mathsfit}{bold}{\encodingdefault}{\sfdefault}{bx}{n}

\usepackage{fontawesome5}
\usepackage{hyperref}
\usepackage{url}
\usepackage{booktabs}
\usepackage{graphicx}
\usepackage{mathtools}
\usepackage{bbm}
\usepackage{amssymb}
\usepackage{mathrsfs}
\usepackage{multirow}
\usepackage{enumitem}
\usepackage{svg}
\usepackage{capt-of}
\usepackage{wrapfig}
\usepackage{caption}
\usepackage{mathrsfs}
\usepackage{euscript}

\newsavebox{\generalizationbox}

\newcommand{\modelname}[1]{{\footnotesize #1}}

\title{\includegraphics[height=1.3em]{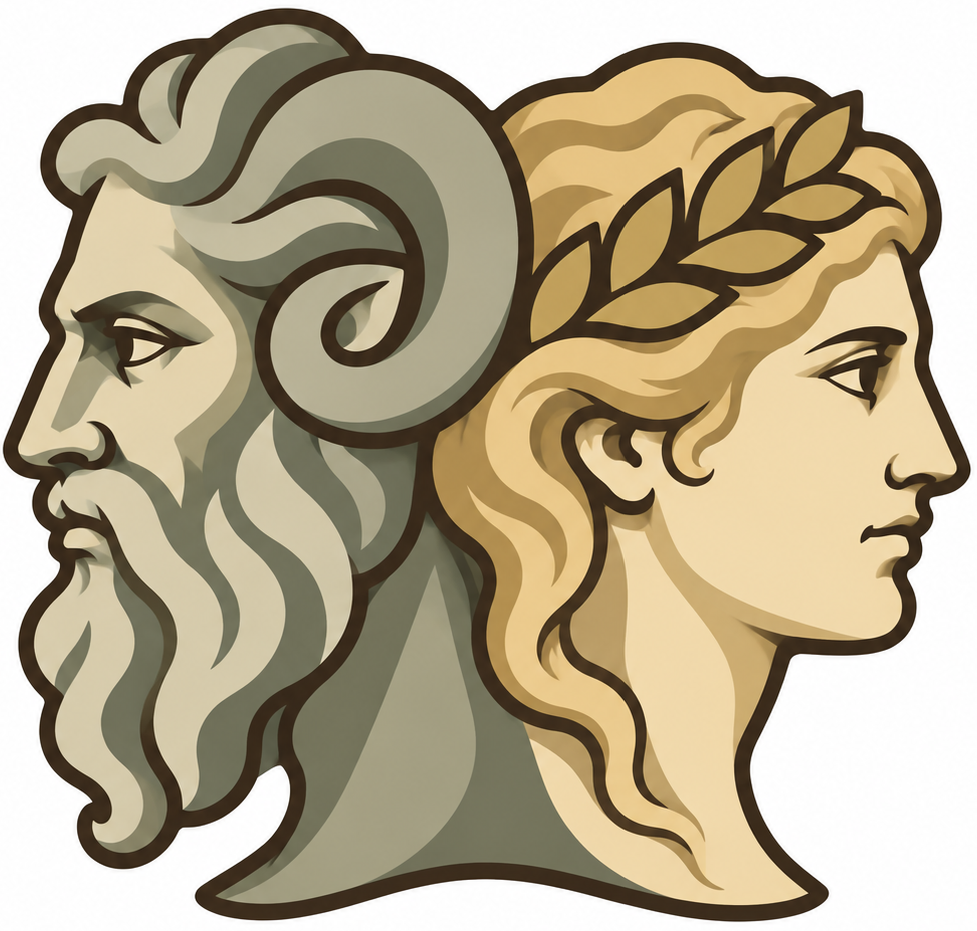}\hspace{0.3em}Janus: \\Enhancing Long-Horizon Autonomous\\ Research with Co-Evolving World Models} 

\author{Mengdi Liu\textsuperscript{1,2}\thanks{Equal Contribution. Mengdi Liu is the project leader.}~, 
Chenghao Jia\textsuperscript{1,2*}, 
Hong Chang\textsuperscript{1,2}\thanks{Corresponding to Hong Chang: \texttt{\textless hanghong@ict.ac.cn\textgreater}}~, 
Mingjie He\textsuperscript{1,2},
Shiguang Shan\textsuperscript{1,2},
Xilin Chen\textsuperscript{1,2}\\
\textsuperscript{1} State Key Laboratory of AI Safety, Institute of Computing Technology, CAS, China\\
\textsuperscript{2} University of Chinese Academy of Sciences (CAS), China\\
}

\iclrfinalcopy 

\begin{document}

\maketitle
\lhead{\textit{2026-10-10}}

\vspace{-2.4em}

\noindent\hspace*{0.4em}
{
  \faGithub\hspace{0.4em}
  \textbf{Project page:}\enspace
  \href{https://github.com/bilemond/JANUS}
  {https://github.com/bilemond/JANUS}
\par}
\vspace{1.4em}

\begin{abstract}
Long-horizon autonomous research has shown promise in accelerating scientific progress through sustained cycles of hypothesis generation, implementation, and validation. However, this \textit{Generate-Execute-Feedback} paradigm relies on reactive trial-and-error, where costly and time-consuming executions introduce severe feedback delays, throttling evolutionary progress under explosively expanding search space. To address this challenge, we propose JANUS, a framework that unifies future outcomes forecasting and past experience reflection through a \textit{Predict-Verify-Evolve} loop driven by a co-evolving world model. Within an evolving graph, the research agent proposes candidates while world model scores them by forecasting value improvement, information gain, and exploration utility, then accordingly selects, refines, or prunes branches prior to execution. After receiving ground-truth feedback, JANUS distills experiences into evolving skills: hierarchical skills for research agent, exploration-strategy skills for world model, and their coordination protocols.
This mechanism enables continual co-evolution: the research agent sharpens candidate generation while the world model refines its predictive judgment over extended horizons.
Extensive evaluations on MLS-Bench-Lite and real research tasks demonstrate that JANUS significantly outperforms state-of-the-art baselines, achieving both higher evolving efficiency and ceiling.

\end{abstract}

\section{Introduction}

Autonomous research (AutoResearch)~\citep{autoresearch,mlreplicate2026,autosci2026, liu2026mindflow0} operationalizes fully automated, end-to-end scientific discovery~\citep{internagent,alphaevolve} through an iterative loop, wherein agents generate candidate solutions, evaluate outcomes, incorporate execution feedback, and use accumulated evidence to guide subsequent iterations.
Such discovery loops have driven substantial progress across diverse domains, including algorithm design~\citep{alphaevolve,romera2024mathematical}, automated model training~\citep{aide, nam2026mle}, open-ended scientific problem solving~\citep{georgiev2025mathematical,anthropic2026riemann}, and systems design~\citep{jaber2026autokernel,cao2026k}.

Recent AutoResearch frameworks have made significant progress by leveraging agentic workflows~\citep{airesearcher, internagent,autoresearchclaw2026, aris2025} or execution-driven heuristic search~\citep{aide, chen2026mars, arbor2026}. These frameworks typically follow a \textit{\textbf{“Generate-Execute-Feedback”}} loop, wherein the system iteratively  refines code based on runtime output~\citep{react}.
However, this paradigm suffers from a severe \textit{execution bottleneck}, as physical execution is computationally expensive and time-consuming, often requiring hours of model training, simulation, and compilation.
Such expensive trial-and-error is severely magnified under \textit{expansive search spaces}: as ongoing experiments continually spawn new hypotheses, the space of possible research trajectories branches combinatorially over time.

Fundamentally, overcoming this bottleneck requires going beyond the incremental gains in conventional iterative refinement. Instead, the core objective of AutoResearch poses a \emph{scaling} problem: \textit{as research compute expands, how should a system navigate long-horizon exploration to maximize cumulative progress?}
We formalize this relationship between research budget and achievable progress as the \textit{\textbf{evolving scaling law}} of autonomous research.
A desirable research system should improve both its \emph{evolving efficiency}---how rapidly progress accumulates—and its \emph{evolving ceiling}---how far continued exploration advances.
To achieve the desired scaling behavior in AutoResearch, we advocate for a paradigm shift from reactive trial-and-error to \textit{\textbf{metacognitive exploration}}: 
(i) proactively forecasting outcomes before physical execution, 
(ii) adaptively allocating execution resources during search, and 
(iii) systematically turning past experiences into reusable research intuition.

\begin{figure*}[tb]
    \centering
	\includegraphics[width=1\textwidth]{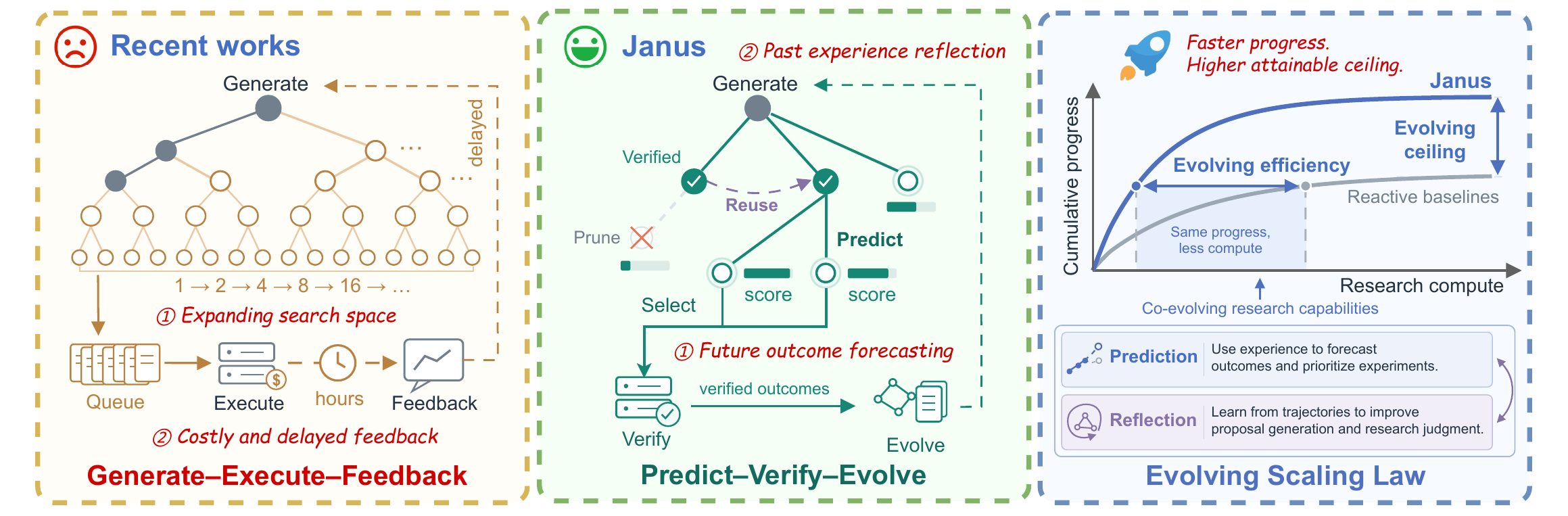}
    \vspace{-0.5cm}
	\caption{\textbf{Conceptual comparison between recent works and JANUS.} 
Unlike traditional \textit{Generate--Execute--Feedback} frameworks that suffer from execution bottlenecks, JANUS introduces a \textit{Predict--Verify--Evolve} loop to forecast outcomes, prune unpromising branches, and reuse past experience. Consequently, JANUS establishes a superior \textit{Evolving Scaling Law}, significantly improving both \textit{evolving efficiency} and the attainable \textit{evolving ceiling} over reactive baselines.
}
	\label{fig: illus}
    \vspace{-0.6cm}
\end{figure*}


To this end, we propose \textbf{JANUS}\footnote{In Roman mythology, JANUS is the two-faced god who looks simultaneously toward the future and the past. Similarly, our framework embodies this duality: one face looks forward as a world model to forecast experimental outcomes prior to execution, while the other looks backward to assimilate ground-truth feedback, driving the co-evolution of the research agent, the world model and their coordination.
}, 
a framework that unifies future outcome forecasting and past experience reflection through a \textbf{\textit{“Predict-Verify-Evolve"}} loop driven by a co-evolving world model.
Specifically, we formalize an \textbf{\textit{evolving research graph}} to serve as the unifying foundation, enabling dynamic branch exploration, structured experience management, and topology-aware performance attribution.
Within this graph, JANUS first \textbf{\textit{forecasts before acting}}: the orchestrator generates candidate proposals to expand the graph, while the world model evaluates them via an adaptive exploration strategy balancing value improvement, information gain, and exploration utility—then selecting, refining, or pruning branches prior to execution.
Upon receiving execution feedback, JANUS then \textbf{\textit{reflects on experience}}, distilling raw execution logs into hierarchical and role-specific evolving skills.
This closed-loop process drives the continual co-evolution of the research agent, the world model, and their coordination protocols, empowering the system to continuously learn and evolve while transcending myopic trial-and-error across complex, non-monotonic research spaces.

Empirically, we comprehensively evaluate JANUS across MLS-Bench-Lite~\cite{lyu2026mls} and real-world research tasks covering algorithm optimization, system design, and scientific discovery.
We demonstrate that JANUS significantly outperforms state-of-the-art baselines, consistently generating valid and competitive solutions across various backbone models.
We further trace its \textit{evolving scaling laws}, showing that JANUS achieves superior evolving efficiency with lower compute budgets and a higher evolving ceiling before plateauing.
Ablation studies further validate the superiority of both the adaptive exploration strategy and experience-driven co-evolution.

In summary, our \textbf{contributions} are as follows: 
\begin{itemize}[leftmargin=2.4em,topsep=1pt]
    \item We propose JANUS, a framework that unifies future outcome forecasting and past experience reflection through a \textit{“Predict-Verify-Evolve”} loop driven by a co-evolving world model.
    \item 
    We introduce an evolving research graph, upon which we design a world-model-guided adaptive exploration strategy to 
    navigate research toward superior trajectories and an experience-driven co-evolution mechanism for the research agent, world model, and their coordination.
    \item We demonstrate that JANUS achieves state-of-the-art performance across MLS-Bench-Lite and real-world research tasks, exhibiting superior \emph{evolving efficiency} and a higher \emph{evolving ceiling} as additional search compute is invested.
\end{itemize}



\section{Related Works}

\vspace{-1mm}
\subsection{Autonomous Research Agent}


\vspace{-1mm}
\paragraph{Pipeline-Based Systems.}
Early systems organize research as a sequence of explicit stages. The AI Scientist~\cite{lu2024aiscientist} established an end-to-end idea-to-paper workflow,
while Agent Laboratory~\cite{schmidgall2025agentlab} structured a similar process around specialized research assistants and human supervision. Subsequent systems incorporate richer tool use and domain-specific capabilities~\cite{cycleresearcher2024,kosmos2025,dolphin2025,codescientist2025,novelseek2025,scientistone2026}.
Such pipelines are modular and interpretable, but they can propagate early errors and offer limited flexibility for revisiting alternative research directions.

\vspace{-0.5em}
\paragraph{Search-Based Systems.}
Search-based research agents advance through explicit solution exploration and refinement. AIDE~\cite{aide} performs iterative code-space search, while MLE-STAR~\cite{nam2026mle} refines targeted pipeline components.
AIRA~\cite{toledo2026ai} studies greedy, MCTS, and evolutionary search policies, and AI Scientist-v2~\cite{yamada2025aiscientistv2} extends tree search to research plans and experiments. ML-Master~\cite{liu2025ml} and MARS~\cite{chen2026mars} further extend exploration through parallel experience reuse and budget-aware search.
These systems rely on costly execution feedback and lack adaptive prioritization of research directions. Janus instead uses an evolving world model to select promising directions before execution.

\vspace{-0.5em}
\paragraph{Multi-Agent Systems.}
Multi-agent systems decompose research across agents with complementary responsibilities. Early systems use specialized roles to structure research~\cite{schmidgall2025agentlab,airesearcher}, while R\&D-Agent~\cite{rdagent} couples researcher and developer agents through experimental feedback. Recent systems further introduce persistent memory and self-evolution~\cite{evoscientist2026}, or adversarial executor--reviewer collaboration for reliable scientific reasoning~\cite{aris2025}. Arbor~\cite{arbor2026} instead uses a hypothesis tree coordinated by a long-lived agent and executed by short-lived agents.
However these systems still face challenges in efficient information exchange and role-specific experience utilization. Janus addresses this through a shared evolving research graph and an experience-driven co-evolution mechanism.

\vspace{-1mm}
\subsection{LLMs as World Models}

\vspace{-1mm}
Recent work suggests that LLMs can function as implicit or explicit world models for planning and decision making. RAP~\cite{hao2023reasoning} frames reasoning as planning with an LLM world model, while complementary approaches induce structured domain models (e.g., PDDL) from language for classical planning~\cite{guan2023leveraging}.
~\cite{gkountouras2025language} integrate LLM agents with causal world models that explicitly capture environment dynamics.
Recent works~\cite{gu2024your,fang2025webevolver,chae2025web,shen2026world} further use LLM-based world models to simulate action outcomes and evaluate candidate trajectories for agentic planning.
Most existing approaches target relatively fixed tasks and do not continually update the world model. Janus instead enables the world model to evolve with the research process, allowing its ability to improve over time.

\section{The JANUS Framework}
\label{sec:method}

\subsection{Problem Formulation}
\label{sec:problem}

\paragraph{Evolving Research Graph.}

As autonomous research unfolds over long horizons, systems suffer from severe \textit{context drift} and intractable \textit{credit assignment} across iterations. To address these challenges, we explicitly maintain an evolving directed acyclic graph (DAG), enabling branch exploration, experience reuse, and topology-aware performance attribution, denoted (at step $t$) as:
\begin{equation}
    \mathcal{G}_t=(\mathcal{V}_t,\mathcal{E}_t), \qquad v_i=(x_i,\hat{V}_i,y_i,e_i)\in\mathcal{V}_t, \qquad \mathcal{E}_t\subseteq\mathcal{V}_t\times\mathcal{V}_t.
    \label{eq:research_graph}
\end{equation}


Each node $v_i$ is partitioned into an \textsc{Open} or \textsc{Closed} state. An \textsc{Open} node records a proposal $x_i$ and its world-model-estimated value $\hat{V}_i$, whereas a \textsc{Closed} node additionally records its implementation $y_i$ and experimental results $e_i$.
An edge $(v_i,v_j)$ indicates that the proposal $x_j$ is obtained by directly modifying $x_i$, extending $x_i$, or combining mechanisms from $x_i$ and other nodes.
As research progresses, the graph dynamically expands with new proposals, continually updating value estimates from the world model and validating with experimental results, thereby characterizing the full lifecycle evolution of autonomous research.

\vspace{-0.5em}
\paragraph{AutoResearch as POMDP.}

The iterative process of AutoResearch can be formulated as a budget-constrained partially observable Markov decision process (budget-constrained POMDP), denoted by
$\mathcal{M}=(\mathcal{S},\mathcal{A},\mathcal{O},
\mathcal{T},\Omega,\mathcal{R})$.
We define states, observations, and actions at step $t$ as: 
\begin{equation}
    s_t=(\mathcal{G}_t,B_t,z_t)\in \mathcal{S}, \qquad o_t=(\mathcal{G}_t,B_t)\in \mathcal{O}, \qquad a_t=(a_t^{\mathrm{orc}},a_t^{\mathrm{exe}})\in \mathcal{A}.
    \label{eq:pomdp_components}
\end{equation}
Here, $\mathcal{G}_t$ is the evolving research graph introduced above, $B_t$ is the remaining budget, and $z_t$ captures unobserved factors such as attainable performance limits, test distribution, and underlying execution mechanisms.
Conditioned on $o_t$, the research policy $\pi$ selects an action comprising proposal generation $a_t^{\mathrm{orc}}$ by orchestrator and the corresponding implementation and execution $a_t^{\mathrm{exe}}$.
The transition model $\mathcal{T}$ and observation model $\Omega$ define $s_{t+1}\sim\mathcal{T}(\cdot\mid s_t,a_t)$ and $o_{t+1}\sim\Omega(\cdot\mid s_{t+1},a_t)$.
We define the immediate reward at step $t$ as the improvement in the best empirical performance so far:
\begin{equation}
    \mathcal{R}_t := \mathcal{R}(s_t,a_t,s_{t+1})
    = \max_{v_i\in\mathcal{V}_{t+1}} r(e_i)
    - \max_{v_i\in\mathcal{V}_t} r(e_i),
    \label{eq:research_reward}
\end{equation}
where $\mathcal{V}_t$ is the node set of the research graph
at step $t$, and $r(e_i)$ denotes the empirical performance
score obtained from experimental results $e_i$, with
larger values indicating better performance.

\vspace{-0.5em}
\paragraph{Research World Model.}
We introduce a research world model to estimate the value of candidate research actions and optimize the research policy to enable value-guided, pre-execution filtering and refining for more effective exploration.
Given the observation $o_t$ and an orchestrator action $a^{\mathrm{orc}}_t$, the research world model $\mathcal{W}_{\phi}$ estimates its \textit{expected cumulative reward}, integrating both the immediate yield of the action and the long-term return from subsequent research under the remaining budget:
\begin{equation}
    \hat{V}_t(o_t,a_t^{\mathrm{orc}})
    =
    \mathbb{E}_{\pi}\!\left[
        \sum_{\tau=t}^{T-1}\mathcal{R}_{\tau}
        \,\middle|\, o_t,a_t^{\mathrm{orc}}
    \right],
    \qquad
    \text{subject to}\quad
    \sum_{\tau=t}^{T-1}\kappa_{\tau}\leq B_t,
    \label{eq:research_value}
\end{equation}
where $\kappa_{\tau}$ is the research cost at step $\tau$
and $T$ is the terminal iteration.
Using this assessment, the world model optimizes candidate actions and guides their selection, shaping the actions decided by research policy toward higher expected final performance.

\begin{figure*}[tb]
    \centering
	\includegraphics[width=1\textwidth]{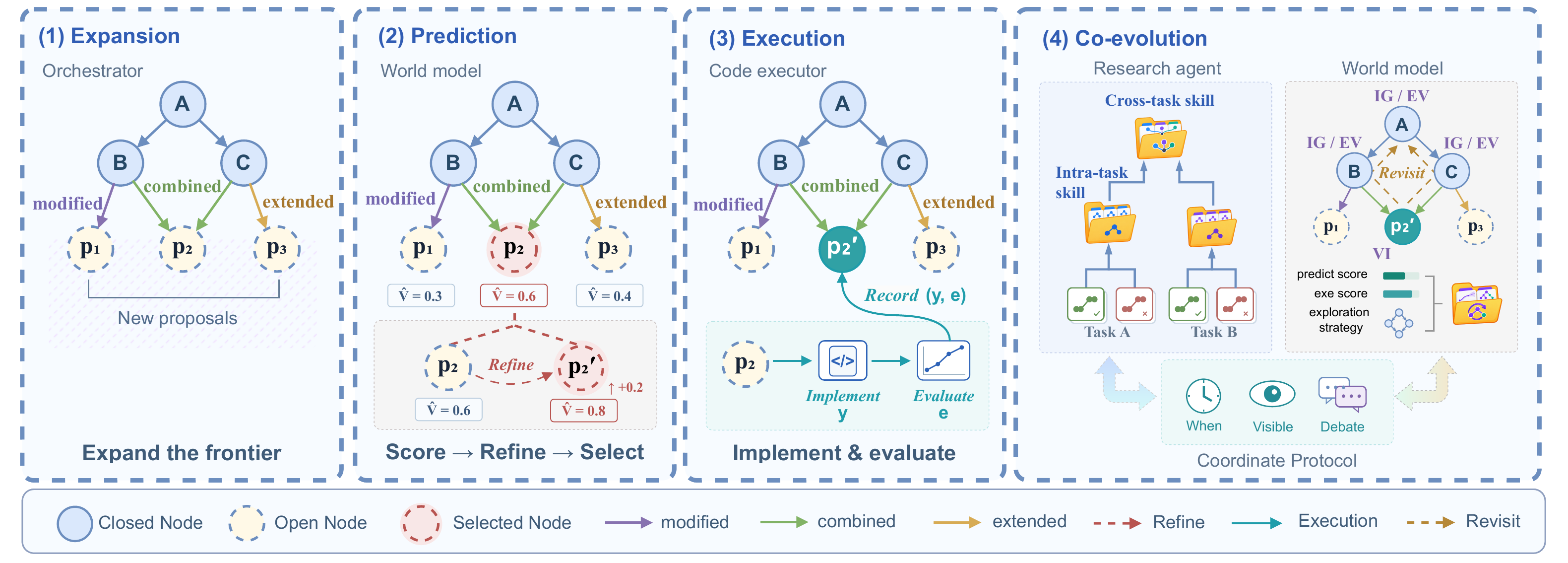}
    \vspace{-0.5cm}
	\caption{The overall framework of JANUS. 
    In each iteration, the orchestrator generates new proposals, the world model scores and refines them to guide selection, and the code executor implements and evaluates them. Experimental feedback drives hierarchical skill abstraction, exploration-strategy reflection, and coordination updates, enabling the research agent and world model to co-evolve.
    }
	\label{fig: method}
    \vspace{-0.5cm}
\end{figure*}

\vspace{-0.25em}
\subsection{The Predict-Verify-Evolve Loop}
\label{sec:Loop}


\vspace{-0.25em}
Building on the evolving research graph, 
JANUS introduces the Predict-Verify-Evolve loop to overcome the high cost and latency of traditional feedback.
As illustrated in Figure \ref{fig: method}, it comprises four key steps: expansion (by orchestrator), prediction and refinement (by world model), implementation and execution (by code executor), and co-evolution.
The details of each step are described as below.

\vspace{-0.5em}
\paragraph{Expansion.}

At iteration $t$, the orchestrator expands the research space by sampling actions $\mathcal{A}_t^{\mathrm{orc}}$, where each action first selects a parent node set $\mathcal{U}_t^{(k)}$ that provides reusable knowledge and then generates a corresponding candidate proposal $x_t^{(k)}$ by modifying, extending, or combining them:

\vspace{-5mm}
\begin{equation}
    \mathcal{A}_t^{\mathrm{orc}}
    =
    \left\{
    a_t^{\mathrm{orc},(k)}
    =
    \left(\mathcal{U}_t^{(k)}, x_t^{(k)}\right)
    \right\}_{k=1}^{N_t},
    \qquad
    a_t^{\mathrm{orc},(k)}
    \sim
    \pi_{\mathrm{orc}}
    \left(
    \cdot
    \mid
    \mathcal{G}_t, B_t, \mathcal{H}_t
    \right),
    \qquad
    \mathcal{U}_t^{(k)} \subseteq \mathcal{V}_t .
    \label{eq:expansion}
\end{equation}
\vspace{-5mm}

The expansion process $\pi_{\mathrm{orc}}$ is conditioned on the current evolving graph $\mathcal{G}_t$, remaining budget $B_t$, and available research skills $\mathcal{H}_t$ (detailed along with its evolution mechanism in Section~\ref{sec:reflection}), and 
$N_t$ denotes the number of generated proposals.
By connecting these proposals to their parents as \textsc{Open} nodes, the research graph systematically expands its exploration frontier.

\vspace{-0.5em}
\paragraph{Prediction and Refinement.}
Drawing on experimental evidence in $\mathcal{G}_t$ and relevant prediction skills from $\mathcal{H}_t$, the world model $\mathcal{W}_{\phi}$ estimates the research value of each candidate proposal $x_t^{(k)}$:
\begin{equation}
\hat{V}_t(x_t^{(k)})=\mathcal{W}_{\phi}\bigl(x_t^{(k)}\mid \mathcal{G}_t,B_t,\mathcal{H}_t\bigr).
\label{eq:prediction}
\end{equation}
This assessment considers both immediate improvement and potential future gains
of candidate proposals within the remaining budget.
Beyond estimating research value, the world model debates with the orchestrator via $\mathcal{D}_{\mathrm{wm}\leftrightarrow\mathrm{orc}}$ to refine proposals $\mathcal{X}_t$ and research values $\hat{V}_t$ when necessary.
Then the world model selects the proposal $x_t^\star$ with the highest estimated research value:

\vspace{-4mm}
\begin{equation}
(\mathcal{X}_t', \hat{V}_t') = \mathcal{D}_{\mathrm{wm}\leftrightarrow\mathrm{orc}}(\mathcal{X}_t; \hat{V}_t), \qquad x_t^\star \in \operatorname*{arg\,max}_{x\in\mathcal{X}_t'} \hat{V}_t'(x).
\label{eq:predictive_selection}
\end{equation}
\vspace{-4mm}

The value estimates are recorded in the corresponding graph nodes, and the selected proposal is passed to the executor.
Section~\ref{sec:wm_operations} details the assessment, refinement, and selection strategy.

\paragraph{Implementation and Execution.}
Given the selected proposal $x_t^\star$, the code executor produces an implementation $y_t$, executes it in the task environment, and updates the remaining budgets:
\begin{equation}
    y_t\sim\pi_{\mathrm{code}}(\cdot\mid x_t^\star,\mathcal{G}_t,B_t,\mathcal{H}_t), \qquad e_t=\operatorname{Eval}(y_t), \qquad B_{t+1}=B_t-\kappa_t .
    \label{eq:execution}
\end{equation}
where $e_t$ denotes the empirical feedback obtained from the environment, and $\kappa_t$ is the total research cost incurred during this iteration.
JANUS records $y_t$ and $e_t$ in the node associated with $x_t^\star$. Therefore, the updated graph $\mathcal{G}_{t+1}$ and budget form the next observation $o_{t+1}$ and state $s_{t+1}$.


\paragraph{Co-evolution.}

JANUS reflects on the updated research graph $\mathcal{G}_{t+1}$, examining how proposal design and code implementation contributed to the experimental outcomes and reassessing the world model's pre-execution judgements in light of this evidence and related experiments.
\begin{equation}
    \mathcal{H}_{t+1}=\operatorname{Evolve}\bigl(\mathcal{H}_t,\mathcal{G}_{t+1}\bigr).
    \label{eq:coevolve}
\end{equation}

Through continual learning from experimental experience, the research agent improves proposal generation, and the world model refines its research judgement, enabling their co-evolution throughout the research process.
The skill evolution mechanism is detailed in Section~\ref{sec:reflection}.

\subsection{World Model Guided Exploration Strategy}
\label{sec:wm_operations}

The core challenge in open-ended research lies in balancing immediate gains against long-term exploration. JANUS leverages its world model to evaluate these complementary value drivers, dynamically adjusting their weights based on recent progress and remaining budget. These priorities govern proposal refinement and selection, steering research exploration trajectory across iterations.


\paragraph{Research Priority Scoring.}
\label{sec:wm_scoring}
The world model evaluates each proposal along three complementary dimensions,
with their formal definitions provided in Appendix~\ref{app:wm_scoring}:
\begin{itemize}[leftmargin=1.4em,topsep=1pt]
    \item \textbf{Value improvement:} expected immediate gain of empirical performance upon execution.
    \item \textbf{Information gain:} 
    expected uncertainty reduction about which proposal should be pursued next.
    \item \textbf{Exploration utility:} expected utility of promising follow-up interventions unlocked by the proposal under the remaining budget.
\end{itemize}
    
    

Rather than assigning absolute scores independently, JANUS evaluates the candidate proposals
$\mathcal{X}_t=\{x_t^{(k)}\}_{k=1}^{N_t}$
through a \textit{\textbf{sparse pairwise tournament}}.
Let $d$ index one of the three research criteria,
$d\in\{\mathrm{VI},\mathrm{IG},\mathrm{EV}\}$,
corresponding to value improvement, information gain, and exploration utility.
For each compared pair $(x_t^{(m)},x_t^{(n)})$, the world model predicts a soft preference:
\begin{equation}
   \rho_{t,d}^{(m,n)}
    =
    P_{\phi}\!\left(
    x_t^{(m)} \succ_d x_t^{(n)}
    \mid
    \mathcal{G}_t,B_t,\mathcal{H}_t
    \right),
    \label{eq:pairwise_pref}
\end{equation}
where $\rho_{t,d}^{(m,n)}$ denotes the predicted probability that
$x_t^{(m)}$ is preferable to $x_t^{(n)}$ under criterion $d$.
To keep the tournament sparse, JANUS adaptively selects informative proposal pairs, prioritizing those with high comparison uncertainty. Let $\mathcal{C}_t$ denote the set of paired indices of proposals.
For each criterion $d$, 
JANUS estimates the latent Bradley-Terry score of $x_t^{(k)}$, denoted as $g_{t,d}(x_t^{(k)})$, by fitting a soft Bradley--Terry model~\cite{bradley1952rank}: 
\begin{equation}
   \operatorname*{min}
    \sum_{(m,n)\in\mathcal{C}_t}
    \mathrm{BCE}\!\left(
        \rho_{t,d}^{(m,n)},
        \sigma\!\left(
            g_{t,d}(x^{(m)})-g_{t,d}(x^{(n)})
        \right)
    \right),
    \label{eq:soft_bt}
\end{equation}
where 
$\sigma(\cdot)$ is the sigmoid function and $\mathrm{BCE}$ denotes binary cross-entropy.
The fitted scores are then normalized to
guarantee $g_{t,d}(x_t^{(k)})\in[0,1]$.

\paragraph{Adaptive Exploration.}
The world model adaptively steers the research exploration trajectory by jointly adjusting research direction and optimizing candidate space.
Given the tournament-derived criterion scores, the research priority of proposal $x_t^{(k)}$ is
\begin{equation}
    \hat{V}_t\!\left(x_t^{(k)}\right)
    =
    \frac{
        g_{t,\mathrm{VI}}\!\left(x_t^{(k)}\right)
        + \beta_t g_{t,\mathrm{IG}}\!\left(x_t^{(k)}\right)
        + \gamma_t g_{t,\mathrm{EV}}\!\left(x_t^{(k)}\right)
    }{
        1+\beta_t+\gamma_t
    },
    \label{eq:research_priority}
\end{equation}
where $\beta_t,\gamma_t\geq0$ are dynamically adjusted according to recent research progress, the remaining budget and the corresponding skills $\mathcal{H}_t$. These priorities further guide the refinement process in Eq.~\ref{eq:predictive_selection}.
The world model maintains a promising research frontier by dynamically pruning low-value proposals according to these priorities.
And when necessary, particularly under sustained stagnation, it engages the orchestrator in debate to refine promising proposals then re-evaluates them.

\subsection{Experience-Driven Co-Evolution}
\label{sec:reflection}
To efficiently leverage sparse feedback from costly experiments and enable cross-task experience reuse, we introduce a hierarchical skill abstraction for research agent, an exploration-strategy reflection mechanism for world model, and an evolving coordination protocol for their communication.

\vspace{-0.5em}
\paragraph{Hierarchical Experience-to-Skill Abstraction.}
To enable structured knowledge transfer, the research agent organizes raw execution logs into a three-level hierarchy: raw experiences ($\mathscr{E}$), intra-task skills ($\mathcal{H}_{\text{intra}}$), and cross-task skills ($\mathcal{H}_{\text{cross}}$).
Specifically, raw successful and failed trajectories $\mathscr{E}_k = \mathcal{T}_k^+ \cup \mathcal{T}_k^-$ are distilled into specialized knowledge (e.g., domain-specific error mitigation) for task $k$, and then generalized into principles (e.g., hypothesis-driven ablation study) across tasks:
\begin{equation}
    \mathscr{E} \xrightarrow{\mathcal{M}_T}\; \mathcal{H}_{\text{intra}} \xrightarrow{\mathcal{M}_G}\mathcal{H}_{\text{cross}}, \ \text{where } \mathcal{H}_{\text{intra}}^{(k)} = \mathcal{M}_T\!\left(\mathcal{T}_k^+ \cup \mathcal{T}_k^-\right), \; \mathcal{H}_{\text{cross}} = \mathcal{M}_G\!\left(\bigcup_k \mathcal{H}_{\text{intra}}^{(k)}\right).
    \label{eq:tri_level_abstraction}
\end{equation}
Here, $\mathcal{M}_T$ and $\mathcal{M}_G$ denote the abstraction-and-update operators for intra-task and cross-task skills.

\paragraph{Exploration-Strategy Reflection Mechanism.}





To mitigate prediction bias and refine exploration guidance, the world model retrospectively examines its prior judgments against accumulated experimental evidence.
Given the execution experience $\mathscr{E}_i$ of proposal $x_i$, JANUS derives prediction skills $\mathcal{H}_{\mathrm{pre}}$ by reflecting on the predicted value improvement of $x_i$ and the information gain and exploration utility of its ancestors $\mathrm{Anc}(x_i)$, and exploration skills $\mathcal{H}_{\mathrm{exp}}$ from recent adaptive exploration strategies across the reflection window $[t',t]$, tracking $\beta_\tau$, $\gamma_\tau$, and the remaining budget $B_\tau$:
\begin{equation}
    \mathcal{H}_{\mathrm{pre}}
    =
    \mathcal{M}_{P}\!\left(
    \mathscr{E}_i,
    g_{\mathrm{VI}}(x_i),
    \left\{g_d(x_j)\right\}_{x_j\in\mathrm{Anc}(x_i)}^{d\in\{\mathrm{IG},\mathrm{EV}\}}
    \right),
    \
    \mathcal{H}_{\mathrm{exp}}
    =
    \mathcal{M}_{E}\!\left(
    \mathcal{G}_t,
    \Big\{
    (\beta_\tau,\gamma_\tau,B_\tau)
    \Big\}_{\tau=t'}^{t}
    \right).
    \label{eq:reflection_mechanism}
\end{equation}
Here $\mathcal{M}_P$ and $\mathcal{M}_E$ denote abstraction-and-update operators for prediction and exploration skills.

\paragraph{Evolving Coordination Protocol.}

To improve collaboration between the research agent and world model,
JANUS further abstracts shared coordination skills
$\mathcal{H}_{\mathrm{coord}}$ from previous debate and refinement.
Specifically, it reflects on when debate was invoked, how the candidate
proposals and their value estimates were revised, and whether these
revisions were supported by subsequent evidence:
\begin{equation}
    \mathcal{H}_{\mathrm{coord}}
    =
    \mathcal{M}_{C}\!\left(
    \left\{
    (\mathcal{G}_\tau,B_\tau),\,
    (\mathcal{X}_\tau,\hat V_\tau)
    \xrightarrow{\mathcal{D}_{\mathrm{wm}\leftrightarrow\mathrm{orc}}}
    (\mathcal{X}_\tau',\hat V_\tau'),\,
    \mathscr{E}_\tau
    \right\}_{\tau=t'}^{t}
    \right).
    \label{eq:coordination_evolution}
\end{equation}
Here, $\mathcal{M}_{C}$ denotes the abstraction-and-update operator for coordination skills, and $[t',t]$ denotes the recent reflection window covering prior
debate, refinement, and their subsequent outcomes.

\begin{table*}[t]
\centering

\caption{
Overall performance on MLS-Bench-Lite.
\textbf{Beat Any}, \textbf{Beat Median}, and \textbf{Beat All} denote the
percentages of tasks that outperform at least one, the median-performing, and all provided human baselines, respectively. Higher values indicate better performance for all six metrics.
}
\label{tab:mls_main}
\vspace{+2mm}
\setlength{\tabcolsep}{4.5pt}
\renewcommand{\arraystretch}{1.18}

\resizebox{1\textwidth}{!}{
\begin{tabular}{llcccccc}

\toprule

\textbf{Method}
& \textbf{Model}
& \textbf{Valid Sub.}
& \textbf{Beat Any}
& \textbf{Beat Median}
& \textbf{Beat All}
& \textbf{Dev Score}
& \textbf{Test Score} \\

\midrule
\addlinespace[1.5pt]

AutoResearch
& \modelname{DeepSeek-V4 Flash}
& 83.3\%
& 76.7\%
& 40.0\%
& 16.7\%
& 26.24
& 25.87 \\

&
\modelname{DeepSeek-V4 Pro}
& 90.0\%
& 83.3\%
& 46.7\%
& 16.7\%
& 27.01
& 27.13 \\


AIDE
& \modelname{DeepSeek-V4 Flash}
& 93.3\%
& 90.0\%
& 53.3\%
& 23.3\%
& 31.12
& 31.11 \\

MLEvolve
& \modelname{DeepSeek-V4 Flash}
& 96.7\%
& 93.3\%
& 56.7\%
& 30.0\%
& 33.89
& 32.99 \\

AIBuildAI
& \modelname{DeepSeek-V4 Flash}
& 90.0\%
& 86.7\%
& 60.0\%
& 33.3\%
& 34.95
& 35.14 \\

Arbor
& \modelname{DeepSeek-V4 Flash}
& 100.0\%
& 96.7\%
& 66.7\%
& 33.3\%
& 38.01
& 37.34 \\


Codex
& \modelname{DeepSeek-V4 Flash}
& 96.7\%
& 96.7\%
& 73.3\%
& 36.7\%
& 39.38
& 38.91 \\

&
\modelname{GPT-5.6 Sol}
& 100.0\%
& 100.0\%
& 86.7\%
& 53.3\%
& 45.23
& 44.81 \\

\midrule

\textbf{JANUS}
& \modelname{DeepSeek-V4 Flash}
& 100.0\%
& 100.0\%
& 76.7\%
& 43.3\%
& 41.99
& 41.07 \\

&
\modelname{GPT-5.6 Sol}
& \underline{100.0\%}
& \underline{100.0\%}
& \underline{90.0\%}
& \underline{56.7\%}
& \underline{47.32}
& \underline{46.88} \\

&
\modelname{Codex (GPT-5.6 Sol)}
& \textbf{100.0\%}
& \textbf{100.0\%}
& \textbf{93.3\%}
& \textbf{63.3\%}
& \textbf{49.96}
& \textbf{49.78} \\

\bottomrule

\end{tabular}
}
\vspace{-1.0em}
\end{table*}

\section{Experiments}

\vspace{-0.25em}
\subsection{Experimental Setup}

\vspace{-0.5em}
\paragraph{Benchmarks.}
We primarily evaluate JANUS on \textit{MLS-Bench-Lite}~\cite{lyu2026mls}, 
comprising 30 research tasks across 12 domains.
Each task requires agents to improve one targeted component of an ML system and demonstrate the improvement generalizes across controlled settings.
To further examine JANUS’s capacity for end-to-end discovery in real-world research, we consider three additional tasks:
nanoGPT training optimization from AutoResearch~\cite{autoresearch},
MinGPT attention kernel optimization from KernelBench~\cite{ouyang2025kernelbench},
and genotype-to-phenotype prediction task from real world.
Together, these tasks span \textit{algorithm optimization, system design, and scientific discovery}.
Detailed task descriptions are provided in Appendix~\ref{app:benchmarks}.

\vspace{-0.5em}
\paragraph{Baselines.}
We compare JANUS with strong baselines including AutoResearch~\cite{autoresearch}, AIDE~\cite{aide}, MLEvolve~\cite{du2026mlevolve}, AIBuildAI~\cite{aibuildai}, Arbor~\cite{arbor2026}, and Codex~\cite{codex}.
Each method is allowed to inspect files, modify code within the permitted scope, and iteratively run experiments until its budget is exhausted.

\vspace{-0.5em}
\paragraph{Metrics.}
We report six metrics on MLS-Bench-Lite.
\textit{Valid Sub.} measures the percentage of tasks whose final submissions successfully complete evaluation and pass validity checks.
\textit{Beat Any}, \textit{Beat Median}, and \textit{Beat All} measure the percentages of tasks that outperform at least one, the median-performing, and all human baselines, respectively.
\textit{Dev Score} and \textit{Test Score}
report the mean normalized task scores under the development and test evaluation settings.
For the three real research tasks, we report validation bits per byte (Val. BPB)
for nanoGPT, speedup over PyTorch eager execution (Speedup) for MinGPT Causal
Attention, and macro-averaged Pearson correlation (Macro $r$) for genotype-to-phenotype prediction.
Detailed metric definitions are provided in Appendix~\ref{app:metrics}.

\vspace{-0.5em}
\paragraph{Implementation details.}
For each run, the agent is allocated a single NVIDIA A100 GPU.
On MLS-Bench-Lite, we allocate a 5-hour wall-clock budget per task for the iterative research process.
We also use the same initial setups, task objectives, and evaluation protocols across all methods to ensure a fair comparison.
We primarily use DeepSeek-V4 Flash as the backbone, supplemented by DeepSeek-V4 Pro and GPT-5.6 Sol.
Further implementation details are provided in Appendix~\ref{app:implementation}.

\vspace{-0.25em}
\subsection{Performance Analysis}

\vspace{-0.25em}
\paragraph{Overall Performance.}

We first evaluate JANUS on MLS-Bench-Lite under controlled protocol, with the primary results reported in Table~\ref{tab:mls_main}.
We highlight the following:
(1) JANUS outperforms all baselines across six evaluation metrics, demonstrating the superior efficacy of our Predict-Verify-Evolve loop in systematically driving meaningful progress.
(2) JANUS achieves 100.0\% valid improvements upon its weakest human baseline and surpasses all provided human baselines on 43.3\% of tasks, establishing a level of performance that directly rivals human expert standards.
(3) JANUS demonstrates robust backbone generality, consistently maintaining its performance advantage across different base model backbones.
Moreover, its modular architecture is fully compatible with existing harnesses; when integrating our loop with Codex, the framework achieves a peak score of 49.78.

\begin{figure*}[t]
    \centering

    \begin{minipage}[t]{0.325\textwidth}
        \centering
        \includegraphics[width=\linewidth]{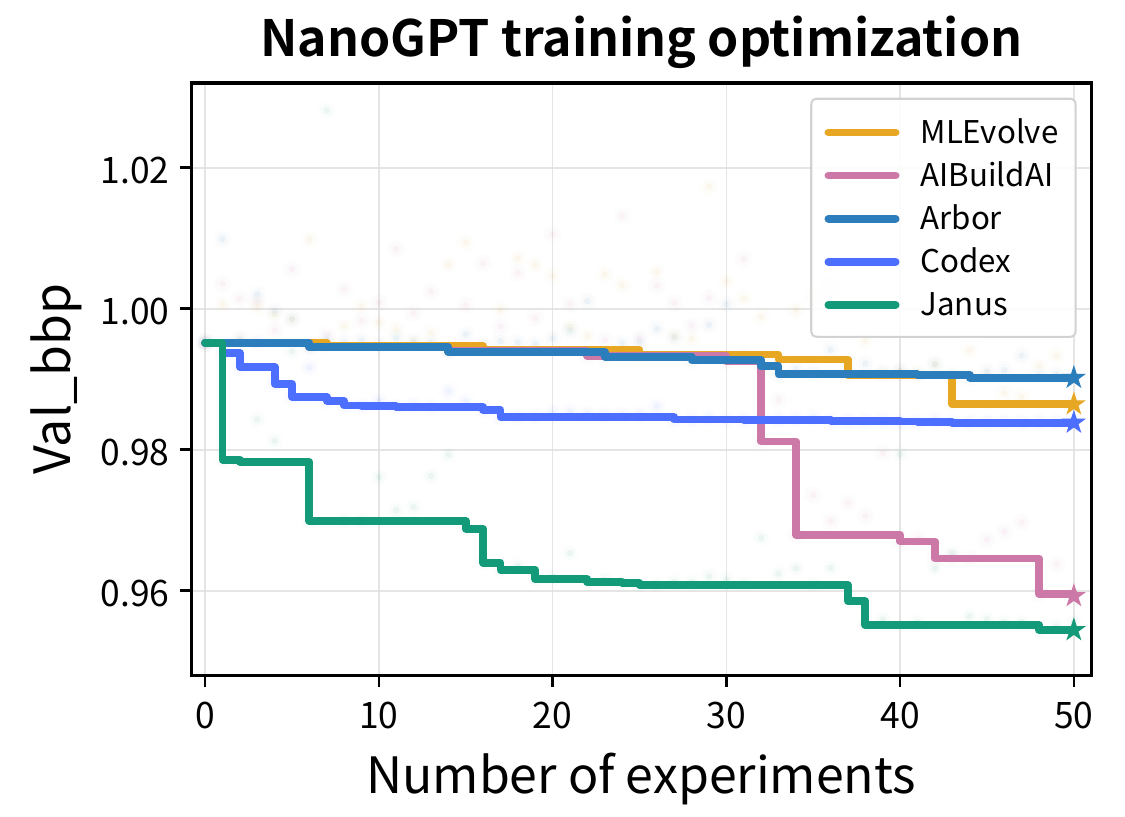}
        \par\vspace{1mm}
    \end{minipage}
    \hfill
    \begin{minipage}[t]{0.325\textwidth}
        \centering
        \includegraphics[width=\linewidth]{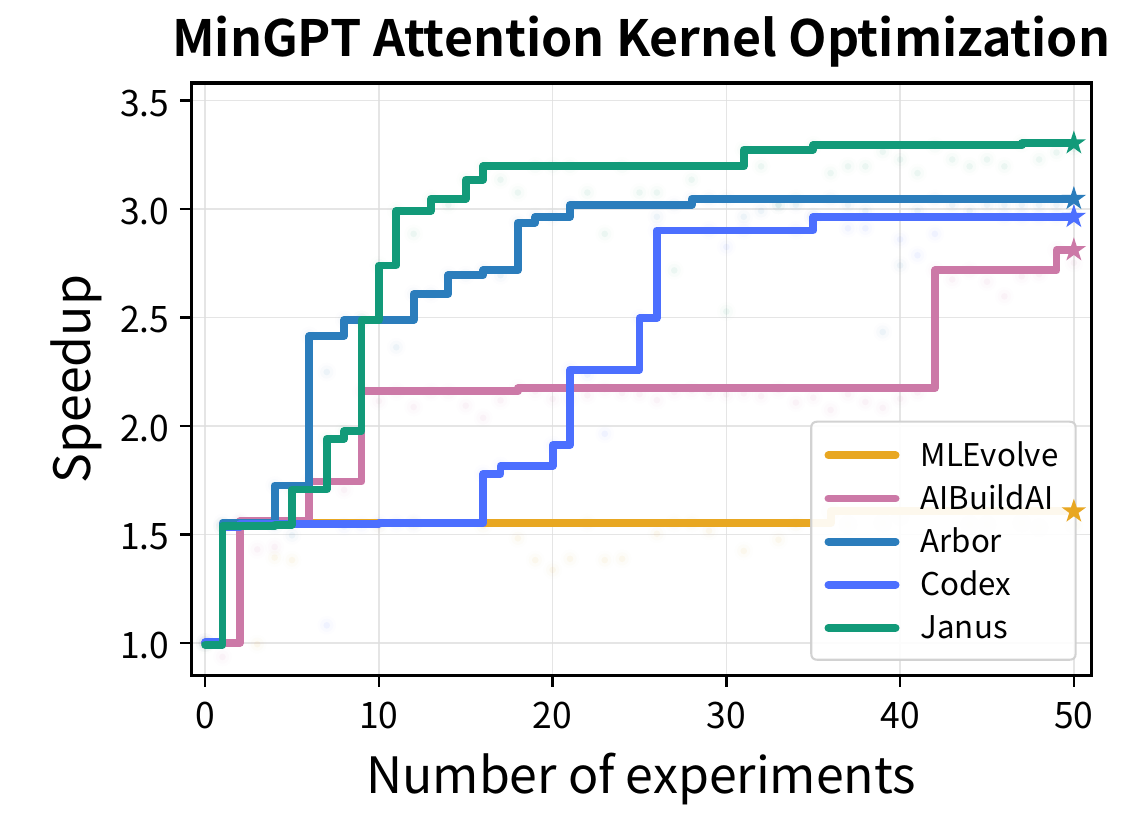}
        \par\vspace{1mm}
    \end{minipage}
    \hfill
    \begin{minipage}[t]{0.325\textwidth}
        \centering
        \includegraphics[width=\linewidth]{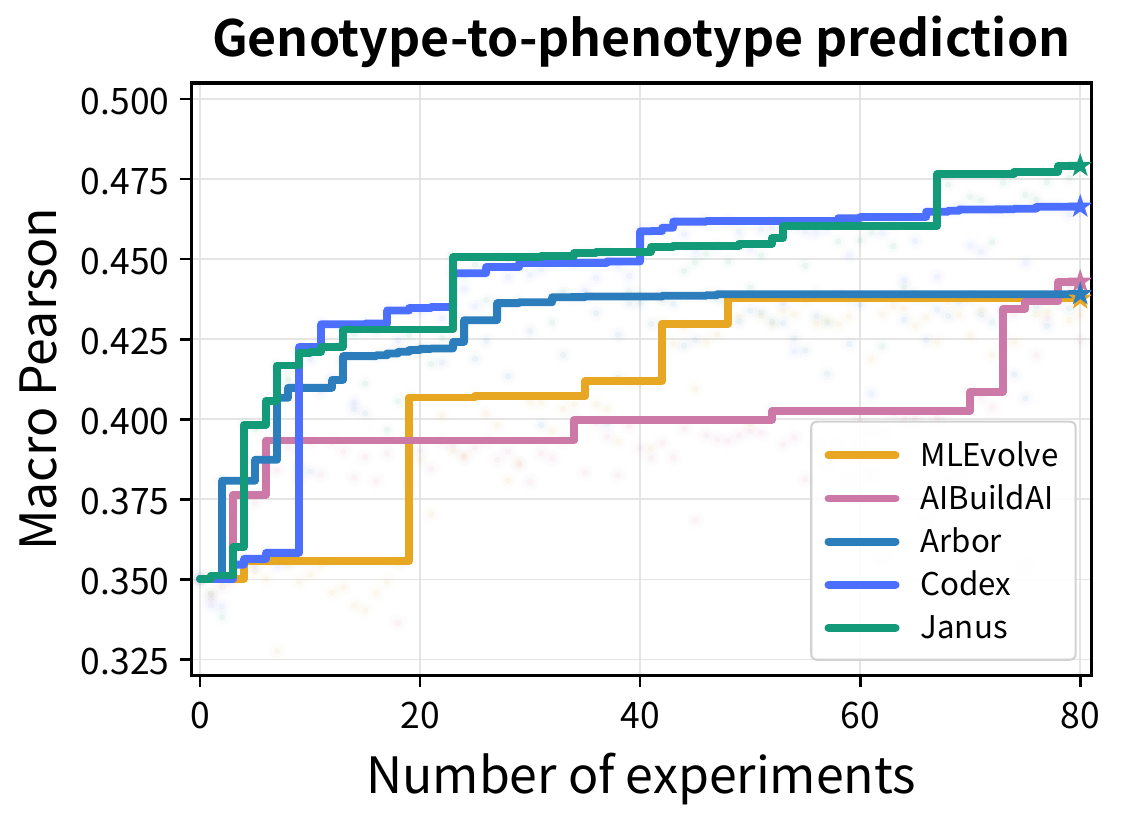}
        \par\vspace{1mm}
    \end{minipage}
    \vspace{-1.0em}
    \caption{
    Research progress (evolving efficiency and ceiling) across three tasks.
    Curves show the best-so-far performance, faint points denote per-experiment scores, and stars mark the final bests.
    }
    \label{fig:research_tasks}
    \vspace{-0.5em}
\end{figure*}

\begin{wraptable}[12]{r}{0.6\textwidth}
\vspace{-1.2em}
\centering
\caption{
Sample-efficiency speedups of JANUS over competing research agents.
Avg.\ and Max.\ denote average and maximum speedups over each baseline.
}
\label{tab:sample_efficiency}

\scriptsize
\setlength{\tabcolsep}{2.5pt}
\renewcommand{\arraystretch}{1.05}
\vspace{-1.0em}

\resizebox{0.6\textwidth}{!}{%
\begin{tabular}{lcccccccc}
\toprule
\multirow{2}{*}{\textbf{Task}}
& \multicolumn{2}{c}{\textbf{MLEvolve}}
& \multicolumn{2}{c}{\textbf{AIBuildAI}}
& \multicolumn{2}{c}{\textbf{Arbor}}
& \multicolumn{2}{c}{\textbf{Codex}} \\
\cmidrule(lr){2-3}
\cmidrule(lr){4-5}
\cmidrule(lr){6-7}
\cmidrule(lr){8-9}
& Avg. & Max.
& Avg. & Max.
& Avg. & Max.
& Avg. & Max. \\
\midrule

NanoGPT
& 35.66 & 43.00
& 16.13 & 34.00
& 26.09 & 44.00
& 7.53  & 43.00 \\

MinGPT
& 1.53 & 7.20
& 2.57 & 4.90
& 1.03 & 2.15
& 2.19 & 3.20 \\

G2P
& 4.15 & 6.33
& 4.30 & 11.67
& 1.18 & 2.09
& 1.51 & 3.00 \\

\midrule
\textbf{Mean}
& \textbf{13.78} & \textbf{18.84}
& \textbf{7.67}  & \textbf{16.86}
& \textbf{9.43}  & \textbf{16.08}
& \textbf{3.74}  & \textbf{16.40} \\

\bottomrule
\end{tabular}
}
\end{wraptable}

\vspace{-0.7em}
\paragraph{Evolving Efficiency and Ceiling.}
We further evaluate JANUS on three real end-to-end research
tasks: nanoGPT training optimization, MinGPT Attention Kernel
Optimization, and genotype-to-phenotype (G2P) prediction.
Figure~\ref{fig:research_tasks} traces the best-so-far performance,
while Table~\ref{tab:sample_efficiency} quantifies
sample-efficiency speedups.
Together, they reveal three key insights.
(1) \textit{Consistent cross-task dominance.}
JANUS leads the best-so-far curves through most of the research
process across all three tasks, while the strongest baseline varies.
(2) \textit{Faster progress per experiment.}
JANUS identifies competitive solutions early in the search,
demonstrating strong evolving efficiency under limited experiment
budgets.
It achieves an average sample-efficiency speedup of $3.74\times$
and a peak of $43\times$ over Codex
(metric details in Appendix~\ref{app:three_task_metrics}).
(3) \textit{Continued improvement beyond temporary plateaus.}
JANUS makes further gains after extended periods without a new
best result, while competing methods plateau earlier.

\begin{figure*}[t]
    \centering
    \includegraphics[width=0.325\textwidth]{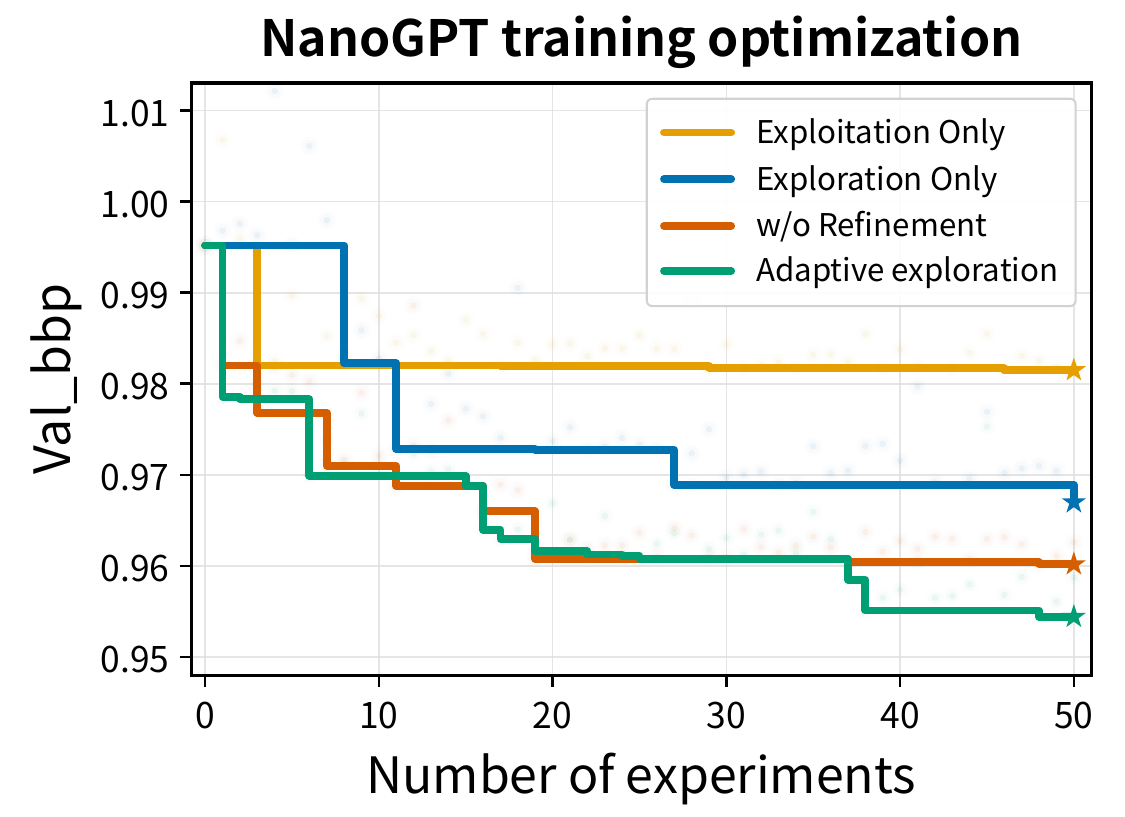}\hfill
    \includegraphics[width=0.325\textwidth]{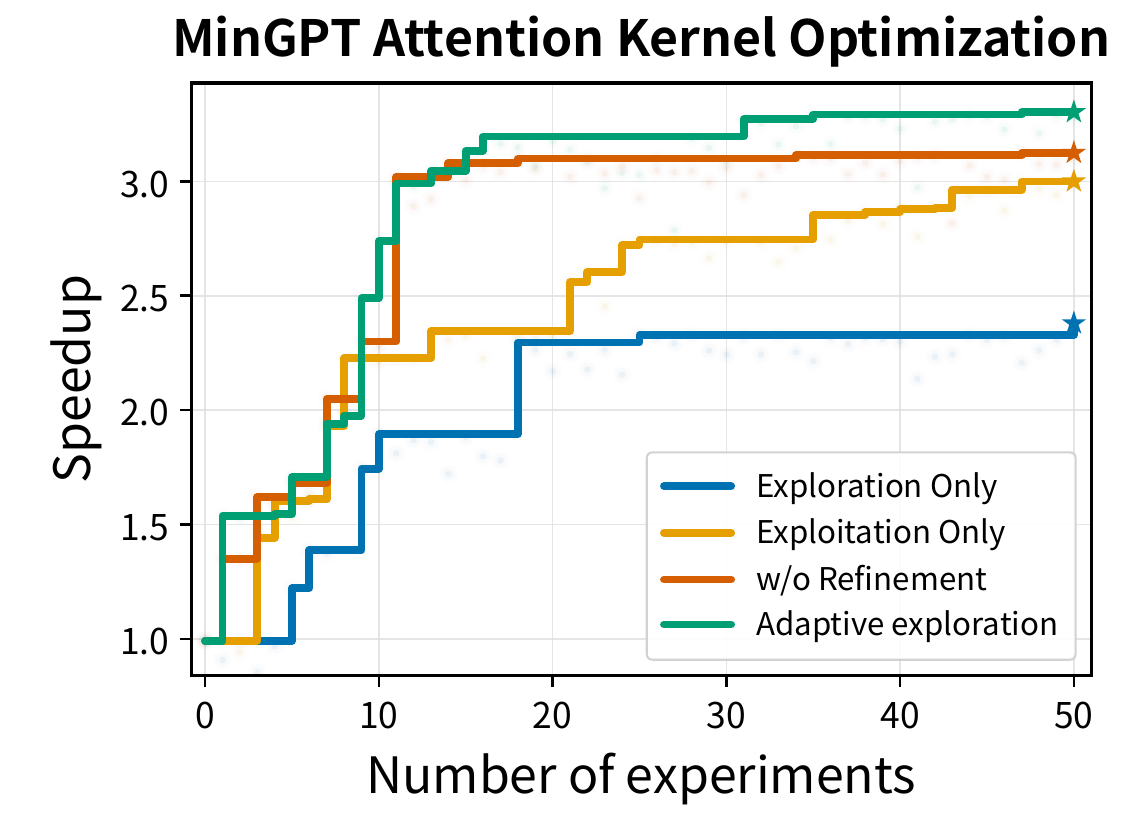}\hfill
    \includegraphics[width=0.325\textwidth]{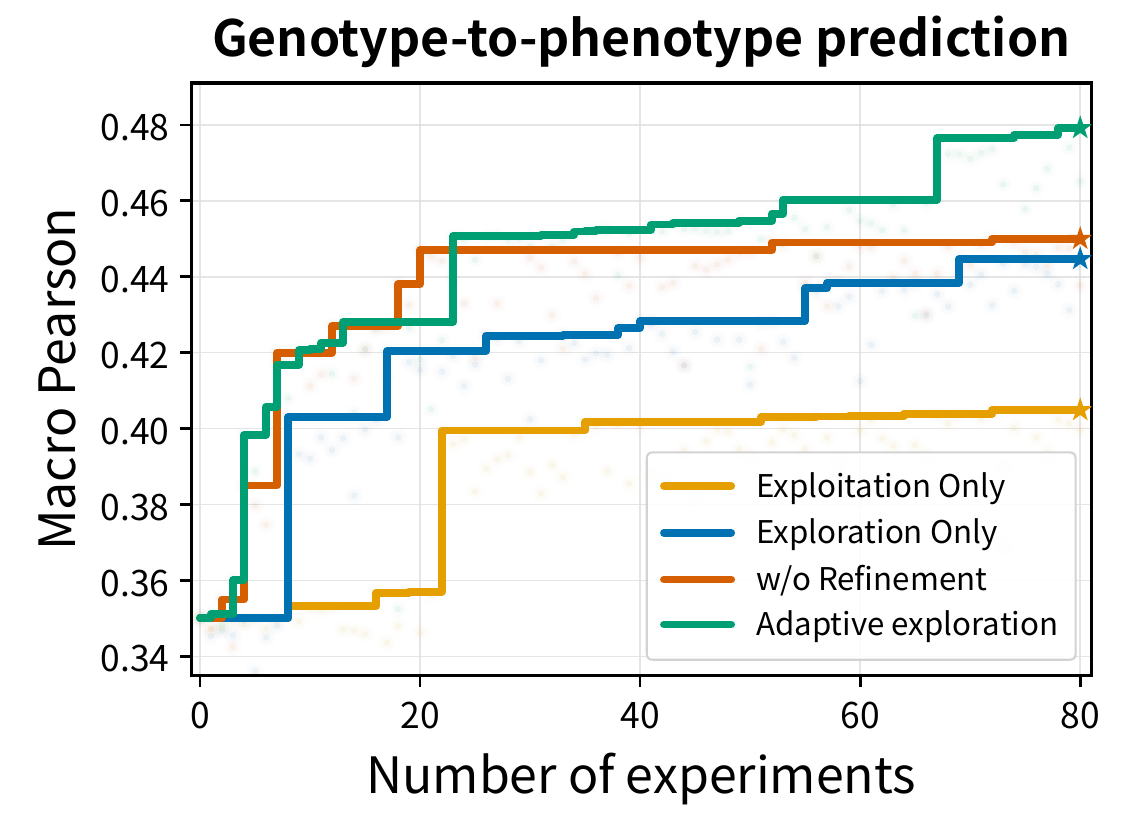}
    
    \vspace{-0.5em}
    \caption{Ablation studies on adaptive exploration strategies across three research tasks.}
    \label{fig:ablation_selector}
    \vspace{-1.5em}
\end{figure*}

\vspace{-0.25em}
\subsection{Ablation Study}

\vspace{-0.5em}
\paragraph{Adaptive exploration strategy.}
We ablate adaptive proposal prioritization and refinement by comparing JANUS with \textit{Exploitation-only}, which prioritizes immediate improvement; \textit{Exploration-only}, which emphasizes exploration utility; and \textit{w/o Refinement}, which disables
world-model-guided refinement of candidate proposals. As shown in Figure~\ref{fig:ablation_selector}, both \textit{Exploitation-only} and \textit{Exploration-only} reduce performance across the three tasks, indicating that a fixed research exploration strategy is insufficient. The \textit{w/o Refinement} variant makes less progress under sustained stagnation, highlighting the importance of revising candidate proposals to optimize the research frontier. Overall, the full framework achieves the strongest final performance across all three tasks, showing that adaptive priority weighting and candidate-space refinement provide complementary benefits.

\begin{figure*}[t]
    \centering
    \includegraphics[width=0.325\textwidth]{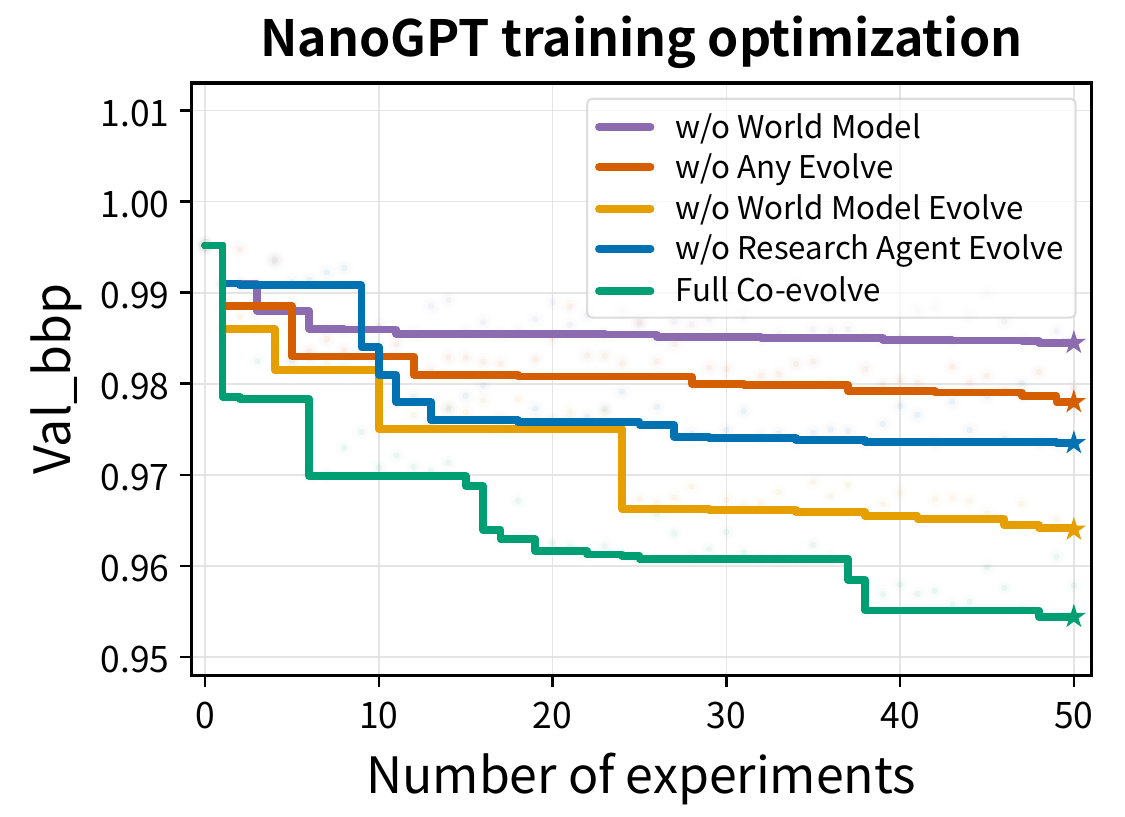}\hfill
    \includegraphics[width=0.325\textwidth]{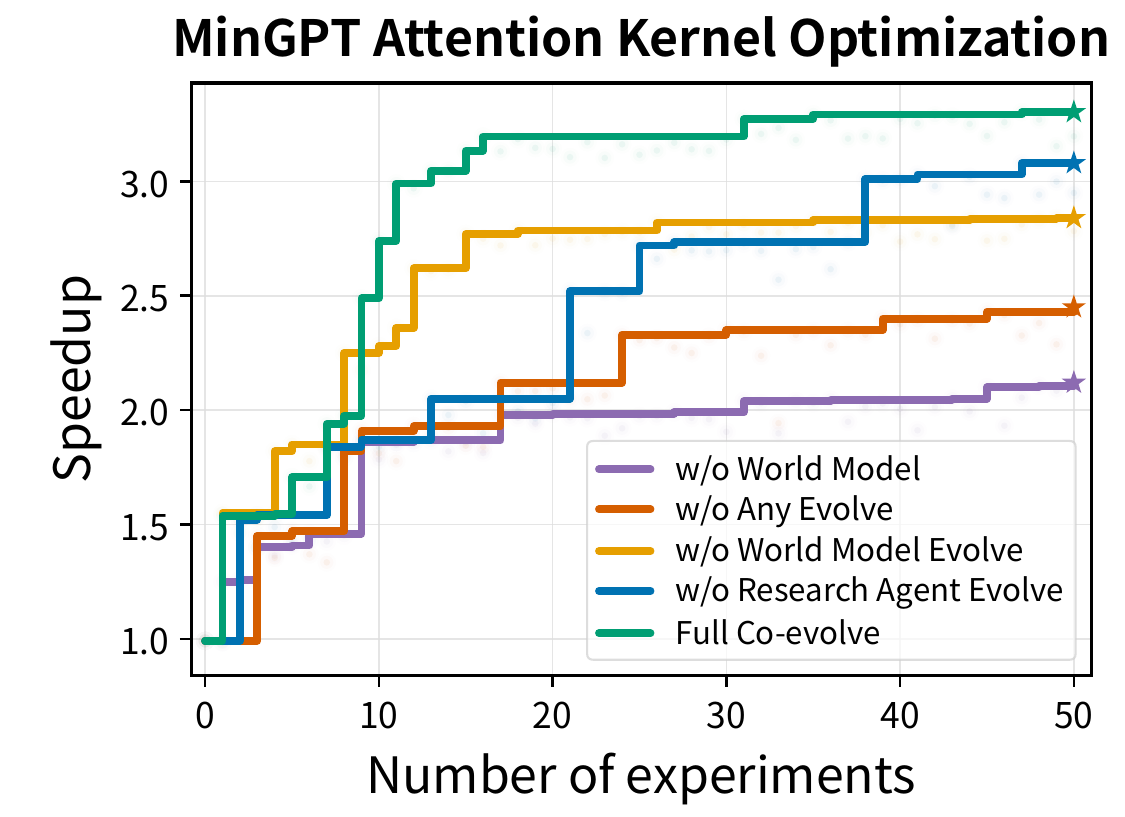}\hfill
    \includegraphics[width=0.325\textwidth]{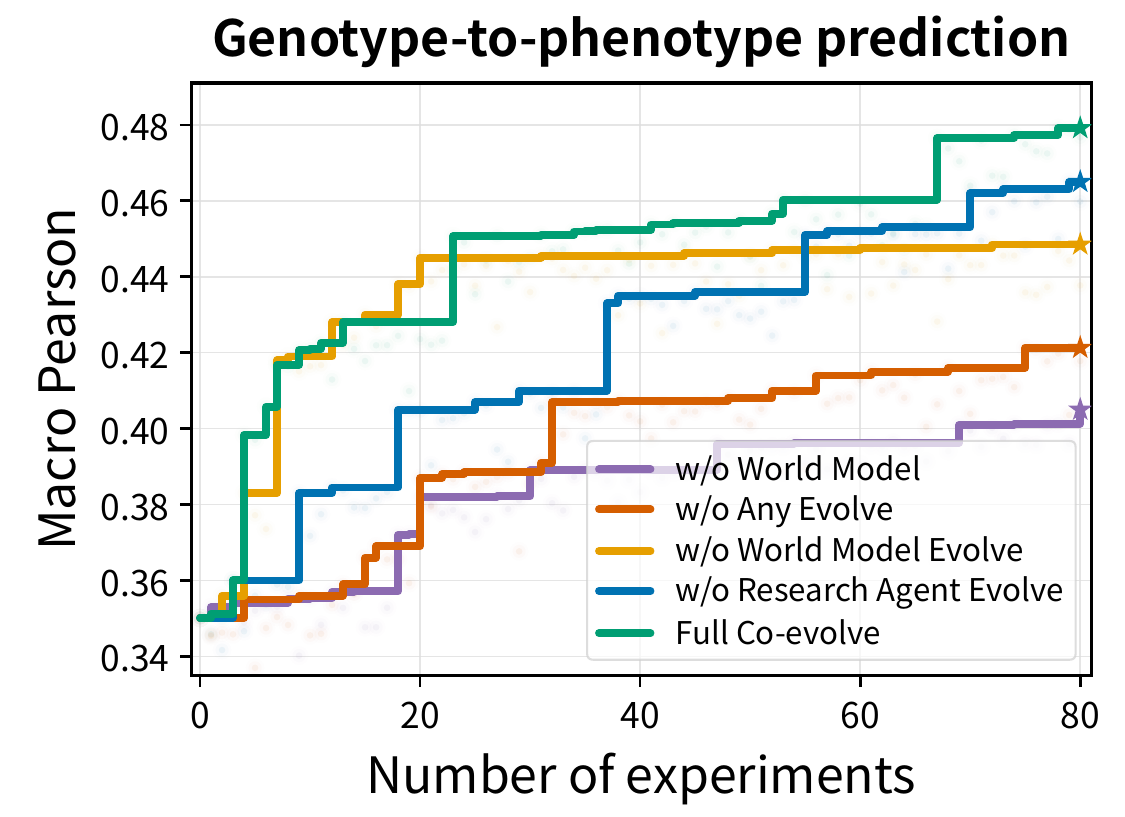}
    
    \vspace{-0.5em}
    \caption{Ablation studies on world model and co-evolution across three research tasks.}
    \label{fig:ablation_coevolution}
    \vspace{-2.5em}
\end{figure*}

\vspace{-0.5em}
\paragraph{World model and co-evolution.}
We further analyze the roles of the world model and co-evolution.
Here, \textit{w/o World Model} removes the world model together with the evolution mechanism, while \textit{w/o “component(s)" Evolve} disables reflection and skill summarization for them.
(1) As shown in Figure~\ref{fig:ablation_coevolution}, removing the world model leads to a substantial performance drop across all tasks, demonstrating the importance of world-model guidance.
(2) Disabling the evolution of either component also degrades performance, indicating that the two evolution processes are both necessary and complementary.
(3) Their effects are also qualitatively different: without world-model evolution, progress is more likely to stall during plateaus, whereas without research-agent evolution, improvement becomes slower throughout the research process.
Together, their co-evolution sustains research progress by continuously turning feedback into better guidance and research behavior.

\begin{figure*}[t]
    \centering

    \begin{minipage}[t]{0.325\textwidth}
        \centering
        \vspace{0pt}
        \parbox[c][0.95\linewidth][c]{\linewidth}{%
            \centering
            \includegraphics[
                width=\linewidth,
                height=0.95\linewidth,
                keepaspectratio
            ]{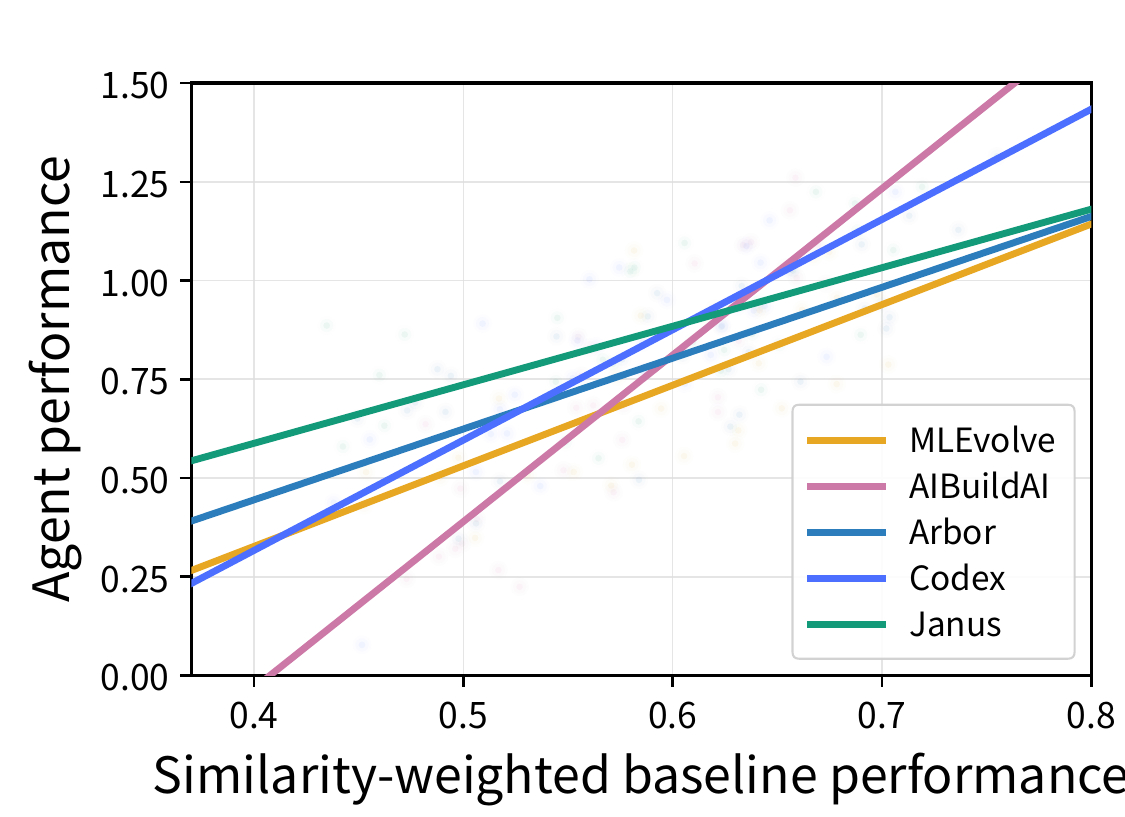}
        }
        \par\vspace{-4mm}
        {\small (a) Beyond baseline imitation}
    \end{minipage}\hfill
    \begin{minipage}[t]{0.325\textwidth}
        \centering
        \vspace{0pt}
        \parbox[c][0.95\linewidth][c]{\linewidth}{%
            \centering
            \includegraphics[
                width=\linewidth,
                height=0.95\linewidth,
                keepaspectratio
            ]{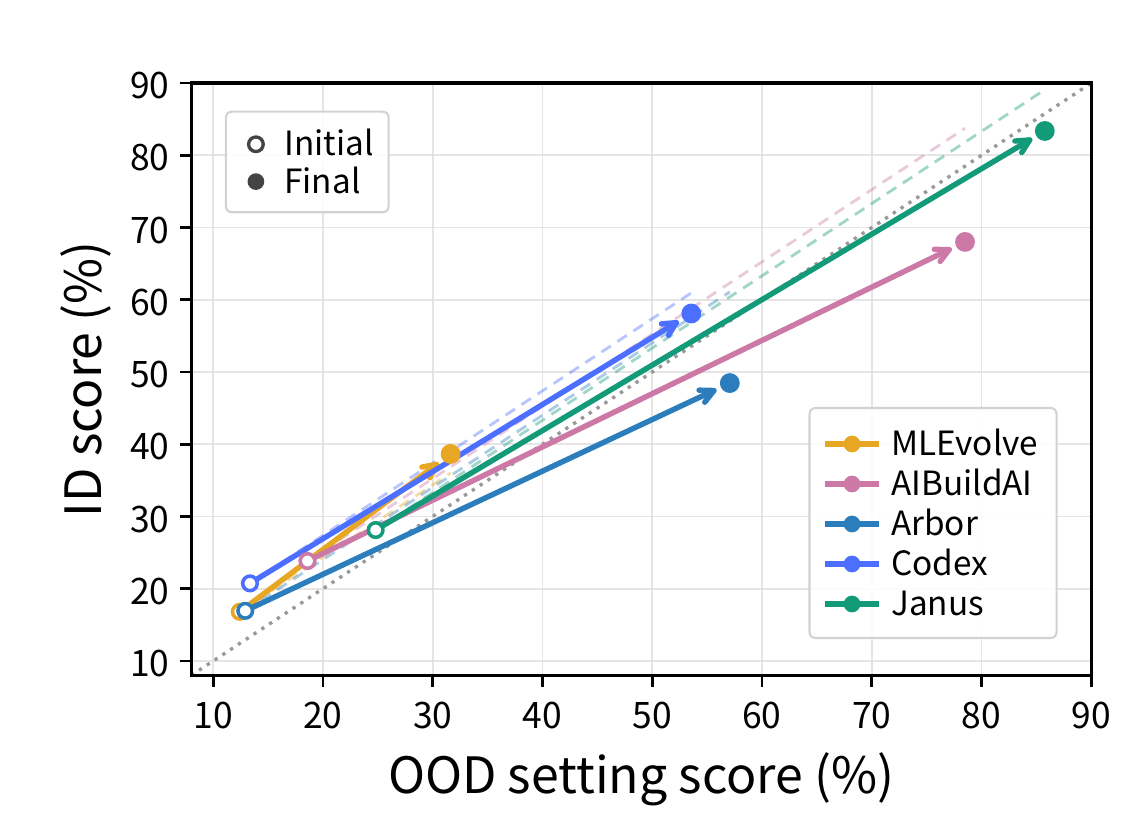}
        }
        \par\vspace{-4mm}
        {\small (b) ID and OOD scores}
    \end{minipage}\hfill
    \begin{minipage}[t]{0.325\textwidth}
        \centering
        \vspace{0pt}
        \parbox[c][0.95\linewidth][c]{\linewidth}{%
            \centering
            \includegraphics[
                width=\linewidth,
                height=0.95\linewidth,
                keepaspectratio
            ]{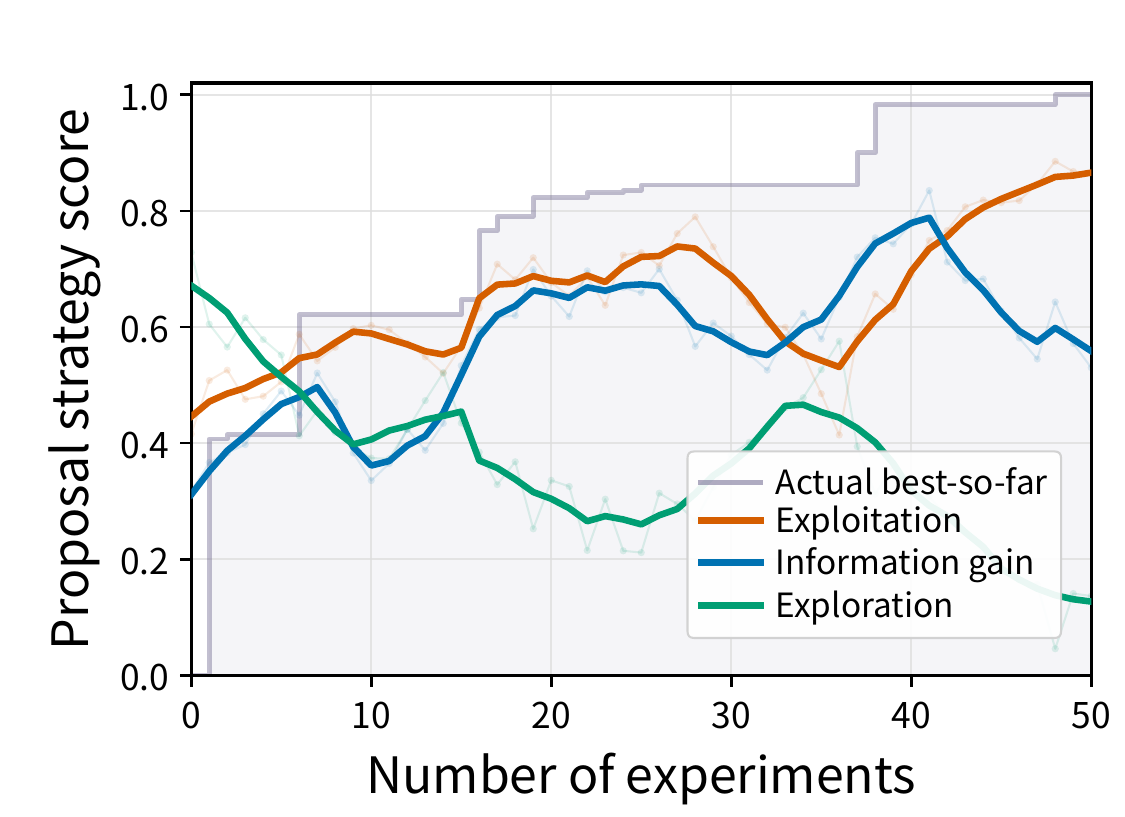}
        }
        \par\vspace{-4mm}
        {\small (c) Dynamic research strategy}
    \end{minipage}

    \vspace{-0.5em}
    \caption{
    \textbf{Framework analysis.}
    (a) Agent performance versus similarity-weighted baseline performance.
    (b) ID and OOD scores of first and final solutions.
    (c) Adaptive research strategy dynamics.
    }
    \vspace{-1.5em}
    \label{fig:framework_analysis}
    \label{fig:reward_similarity}
\end{figure*}

\vspace{-0.25em}
\subsection{Framework Analysis}

\vspace{-0.5em}
\paragraph{Beyond baseline imitation.}
Following MLS-Bench-Lite~\cite{lyu2026mls}, we examine whether research gains primarily arise from imitating strong baselines. As shown in Figure~\ref{fig:reward_similarity}(a),
JANUS maintains strong performance even at low similarity-weighted baseline quality, suggesting that its gains extend beyond direct baseline imitation (details shown in Appendix~\ref{app:baseline_imitation}).

\vspace{-0.5em}
\paragraph{Domain generalization.}
We evaluate whether research gains transfer to out-of-distribution
(OOD) settings across the three tasks. As shown in
Figure~\ref{fig:framework_analysis}(b), JANUS achieves the highest final scores on ID and OOD settings,
while maintaining the smallest ID--OOD gap. This suggests that its
research gains generalize beyond the original evaluation distribution
(further details shown in Appendix~\ref{app:ood}).

\vspace{-0.5em}
\paragraph{Dynamic research strategy.}
We analyze how the research strategy evolves throughout the nanoGPT research process. We use GPT-5.6 Sol to score each proposal along three dimensions---exploitation, information gain, and exploration (details shown in Appendix~\ref{app:strategy_analysis}).
Figure~\ref{fig:framework_analysis}(c) shows four patterns.
\textbf{(1)} Exploration is emphasized early, while exploitation
gradually increases as promising directions emerge.
\textbf{(2)} When performance plateaus, exploration rises and
exploitation declines, followed by improvement in
best-so-far performance.
\textbf{(3)} Increases in information gain are followed by major performance breakthroughs, indicating that resolving uncertainty helps drive subsequent progress.
\textbf{(4)} As the remaining budget decreases, exploitation increases
while exploration declines.
Together, these patterns show a dynamic research strategy throughout the research progress.

\vspace{-0.6em}
\section{Conclusion}

\vspace{-0.6em}
We present JANUS, a framework for long-horizon autonomous research that unifies future outcome forecasting and past experience reflection through a Predict-Verify-Evolve loop. 
Over an evolving research graph, the world model guides proposal refinement and selection by balancing value improvement, information gain, and exploration utility.
JANUS distills experimental feedback into research agent, world-model, and coordination skills, driving system-wide co-evolution.
Evaluations on MLS-Bench-Lite and three real-world research tasks demonstrate that JANUS outperforms strong baselines, achieving greater evolving efficiency and a higher evolving ceiling.

\subsection*{AI use statement}

We used large language models to assist with writing and polishing the manuscript. This assistance included refining sentence structure,
correcting grammatical errors, improving readability, and strengthening
the coherence of the presentation. We reviewed and revised all
AI-assisted text to ensure that it accurately reflected our intended
meaning, technical details, and research findings.

We also used large language models to support literature retrieval
and discovery, including suggesting search terms and identifying
potentially relevant studies. The suggested references served as
starting points for further investigation. We consulted the original
sources, verified bibliographic information, and independently
assessed the relevance and accuracy of the cited material. We retain
full responsibility for the final manuscript, including its claims,
interpretations, and references.



\bibliography{iclr2027_conference}
\bibliographystyle{iclr2027_conference}

\appendix
\clearpage
\appendix





\section{Formal Definitions of Research Priority Criteria}
\label{app:wm_scoring}

This section provides the formal definitions of the three research
criteria introduced in Section~\ref{sec:wm_scoring}.
Let
\[
    \mathcal{I}_t
    =
    \left(
        \mathcal{G}_t,
        B_t,
        \mathcal{H}_t
    \right)
\]
denote the current research context.
For a candidate proposal $x\in\mathcal{X}_t$, the world model predicts
possible empirical feedback according to
$p_{\phi}(e\mid\mathcal{I}_t,x)$, where $e$ denotes the feedback that
would be returned by the environment after implementing and evaluating $x$.

\paragraph{Value improvement.}
Let $R^\star(\mathcal{G})$ denote the best empirical performance among all evaluated solutions contained in research graph $\mathcal{G}$.
Value improvement captures the expected increase in the best empirical
performance after executing a proposal:
\begin{equation}
    \mathrm{VI}_t(x)
    =
    \mathbb{E}_{e\sim p_{\phi}(\cdot\mid\mathcal{I}_t,x)}
    \left[
        R^\star\!\left(
            \hat{\mathcal{G}}_{t+1}^{(x,e)}
        \right)
        -
        R^\star(\mathcal{G}_t)
    \right],
    \label{eq:app_vi}
\end{equation}
where $\hat{\mathcal{G}}_{t+1}^{(x,e)}$ denotes the hypothetical
successor graph after executing $x$ and receiving feedback $e$.
This criterion favors proposals that are likely to directly improve
the current best solution.

\paragraph{Information gain.}
A proposal can be valuable even without directly improving the incumbent
if its outcome helps resolve uncertainty that affects subsequent research.
Let $Z_t$ denote the research direction that should be pursued next under
the world model's current belief.
We define
\begin{equation}
    \mathrm{IG}_t(x)
    =
    \mathrm{H}\!\left(
        Z_t\mid\mathcal{I}_t
    \right)
    -
    \mathbb{E}_{e\sim p_{\phi}(\cdot\mid\mathcal{I}_t,x)}
    \left[
        \mathrm{H}\!\left(
            Z_t\mid\mathcal{I}_t,x,e
        \right)
    \right],
    \label{eq:app_ig}
\end{equation}
where $\mathrm{H}(\cdot)$ denotes entropy under the world model's belief.
For instance, a controlled experiment may provide little immediate
performance gain but distinguish between competing hypotheses, thereby
clarifying which mechanism or research direction should be pursued next.

\paragraph{Exploration utility.}
Beyond clarifying existing choices, a proposal may create new opportunities
for subsequent research.
Let $\mathcal{D}_t(x)$ denote the follow-up proposals already accessible
after selecting $x$, and let $\mathcal{N}_t(x,e)$ denote additional
proposals enabled by the feedback obtained from executing $x$.
We define
\begin{equation}
    \mathrm{EV}_t(x)
    =
    \mathbb{E}_{e\sim p_{\phi}(\cdot\mid\mathcal{I}_t,x)}
    \left[
        Q\!\left(
            \mathcal{D}_t(x)\cup\mathcal{N}_t(x,e);
            \mathcal{I}_t^{x,e}
        \right)
        -
        Q\!\left(
            \mathcal{D}_t(x);
            \mathcal{I}_t^{x,e}
        \right)
    \right],
    \label{eq:app_ev}
\end{equation}
where $\mathcal{I}_t^{x,e}$ denotes the hypothetical research context
after incorporating the feedback from $x$ and deducting its resource cost,
and $Q(\mathcal{D};\mathcal{I})$ denotes the estimated future improvement
attainable by pursuing proposals in $\mathcal{D}$ under the remaining
budget specified by $\mathcal{I}$.
For instance, an experiment may offer limited immediate improvement but
validate a new mechanism, representation, or implementation strategy that
enables a broader family of promising follow-up interventions.

The three criteria therefore capture complementary sources of research
value: direct performance improvement, reduction of uncertainty over
future decisions, and expansion of the subsequent research space.
The world model uses these formal definitions as the scoring criteria for assessing each candidate proposal along value improvement, information gain, and exploration utility.
They provide the semantic basis for the soft pairwise judgments used by
the tournament-based scoring mechanism in Section~\ref{sec:wm_scoring}.

\section{Baselines}
\label{app:baselines}

We compare our framework with AutoResearch~\cite{autoresearch},
AIDE~\cite{aide}, MLEvolve~\cite{du2026mlevolve},
AIBuildAI~\cite{aibuildai}, Arbor~\cite{arbor2026}, and
Codex~\cite{codex}, using the configurations reported in the main text.
For a fair comparison, methods evaluated on the same task receive identical
initial files, task objectives, evaluator access, and resource budgets. Each
method may inspect the provided workspace, modify code within the permitted
scope, and execute experiments iteratively until its budget is exhausted.
When comparing research frameworks under a common backbone, we use the same
underlying language model so that performance differences can be attributed
to the framework rather than model capability.

The original implementations of these systems were developed for different
execution environments and task interfaces. We therefore adapt their public
codebases to both MLS-Bench-Lite and our three research tasks. These adaptations
are restricted to task-facing infrastructure, including input/output adapters,
workspace initialization, evaluator bindings, dependency wrappers, and stopping
conditions. We preserve each method's original research-loop design, agent-role
decomposition, proposal and revision procedures, memory mechanism, and candidate
selection policy. In particular, we do not introduce components from our
framework, task-specific heuristics, or additional scientific knowledge into
the baselines. Thus, the adapted implementations retain the framework design
and algorithmic ideas of the original methods while exposing all systems to a
common experimental interface.

\section{Benchmarks}
\label{app:benchmarks}

\subsection{Three real research tasks}

We additionally evaluate the methods on three executable research problems
covering language-model training, GPU-system optimization, and scientific
prediction.

For \textit{nanoGPT training optimization}, adapted from
AutoResearch~\cite{autoresearch}, the agent improves a compact single-GPU GPT
pretraining system. The editable file contains the model architecture,
attention and MLP blocks, optimizer construction, learning-rate schedules,
batching strategy, and training loop, while the data preparation, tokenizer,
validation split, and metric implementation remain fixed. Training uses a
BPE-tokenized text corpus and is evaluated on a pinned validation shard. The
objective is to obtain the lowest validation bits per byte (BPB) within a fixed
training-time budget. BPB measures the negative log-likelihood assigned to the
validation text per byte rather than per token, making comparisons less
sensitive to tokenization and vocabulary size.

For \textit{MinGPT Causal Attention optimization}, we use KernelBench Level~3,
Task~43~\cite{ouyang2025kernelbench}. The reference operator is a complete
MinGPT masked multi-head self-attention layer. It applies a joint linear
projection to construct queries, keys, and values, reshapes them into attention
heads, computes scaled query--key products, applies a lower-triangular causal
mask and softmax, aggregates the values, and finally applies an output
projection. The agent must implement a drop-in \texttt{ModelNew} with the same
constructor and forward semantics. Candidate implementations may replace the
PyTorch computation with fused PyTorch operators, compilation strategies,
Triton kernels, or custom CUDA code, but must preserve the numerical output.
The objective is to maximize execution speedup over the PyTorch eager reference
after passing all correctness checks.

For \textit{genotype-to-phenotype (G2P) prediction}, the input for a maize
hybrid consists of its SNP dosage sequence together with the environmental
conditions under which it is grown. The benchmark contains 4,269 hybrids,
7,577 MAF/LD-pruned SNP markers, and 185 site--year environments. Environmental
inputs are daily time series describing factors such as mean, maximum, and
minimum temperature, dew-point temperature, solar and photosynthetically active
radiation, precipitation, evapotranspiration, humidity, soil wetness, frost,
wind speed, and vapor-pressure deficit. A prediction is made for each
hybrid--environment pair, allowing the model to capture genetic effects,
environmental effects, and genotype--environment interactions.

The output covers eight agronomic traits: plant height, ear height, grain
moisture, yield in Mg/ha, yield per plant, root lodging, stalk lodging, and test
weight. Equivalently, for a hybrid--environment input $(g,e)$, the prediction
can be written as
\[
    f(g,e)
    =
    \bigl(
    \widehat{y}^{(1)},\ldots,\widehat{y}^{(8)}
    \bigr).
\]
Because the traits have different physical units and ranges, performance is
computed independently for each trait and then macro-averaged. This prevents
traits with more observations or larger numerical scales from dominating the
final score.

\subsection{MLS-Bench-Lite.}
Our main benchmark evaluation uses MLS-Bench-Lite~\cite{lyu2026mls}, a
challenging 30-task subset of MLS-Bench covering all 12 research domains.
The domains include language models, reinforcement learning, robotics, vision
and generation, deep learning, ML systems and efficient ML, AI for science,
time-series forecasting, structured and causal reasoning, trustworthy
learning, classical and adaptive learning, and optimization and theory.
Representative problems include pretraining-optimizer design, diffusion-model
sampling, robot-policy learning, post-training quantization, sparse attention,
time-series imputation, causal discovery, clustering, privacy-preserving
training, and stochastic optimization.

Each task instantiates a community-recognized research question as an executable
problem. An agent receives a task description, a controlled code scaffold,
strong reproduced baselines, designated editable regions, and an evaluation
interface. Training protocols and evaluators are protected, and tasks involving
model components additionally impose capacity constraints. A submission is
evaluated over multiple datasets, environments, model scales, or random seeds,
rather than on a single development instance. The benchmark therefore measures
whether an agent can produce an improvement in the targeted algorithmic
component that remains effective across the prescribed evaluation settings.

\section{Metrics}
\label{app:metrics}

\subsection{MLS-Bench-Lite Metrics}

\paragraph{Score normalization.}
We follow the normalization and within-task aggregation protocol of
MLS-Bench~\cite{lyu2026mls}.
Each raw metric is first transformed into a higher-is-better quantity
$z$ using the task-specified preprocessing and metric direction.
Let $z_{\mathrm{floor}}$ and $z_{\mathrm{ref}}$ denote the corresponding
worst and best reproduced baseline values.
The normalization assigns zero to the worst-baseline anchor and
calibrates the best-baseline anchor to $0.5$.

For a metric with a finite theoretical upper bound
$z_{\mathrm{bound}}$, the normalized score is
\[
    s(z)
    =
    \left[
        \operatorname{clip}
        \left(
            \frac{z-z_{\mathrm{floor}}}
                 {z_{\mathrm{bound}}-z_{\mathrm{floor}}},
            0,1
        \right)
    \right]^{\gamma},
    \qquad
    \gamma
    =
    \frac{\log(0.5)}
         {\log\left(
             \frac{z_{\mathrm{ref}}-z_{\mathrm{floor}}}
                  {z_{\mathrm{bound}}-z_{\mathrm{floor}}}
         \right)}.
\]
For metrics without a finite theoretical upper bound, the normalized
score uses a calibrated sigmoid:
\[
    s(z)
    =
    \max\left\{
        0,\,
        2\sigma\left(
            \frac{z-z_{\mathrm{floor}}}{\lambda}
        \right)-1
    \right\},
    \qquad
    \lambda
    =
    \frac{z_{\mathrm{ref}}-z_{\mathrm{floor}}}{\log 3},
\]
where $\sigma$ is the logistic sigmoid function.

\paragraph{Within-task aggregation.}
For task $t$, evaluation setting $k$ contains normalized metric scores
$s_{t,k,i}$ with task-specified weights $w_{t,k,i}$.
The setting score is their weighted arithmetic mean:
\[
    S_{t,k}
    =
    \frac{\sum_i w_{t,k,i}s_{t,k,i}}
         {\sum_i w_{t,k,i}}.
\]
The task score is obtained by taking the geometric mean across its
$K_t$ evaluation settings:
\[
    S_t
    =
    \left(
        \prod_{k=1}^{K_t} S_{t,k}
    \right)^{1/K_t}.
\]
This aggregation penalizes uneven performance across settings.
Invalid submissions receive a task score of zero.

\paragraph{Development and test scores.}
We construct development and test splits for our experiments.
\textbf{Dev Score} and \textbf{Test Score} evaluate the final submission
on these respective splits using the normalization and within-task
aggregation described above.
Let $S_{t,\mathrm{dev}}^{(m)}$ and $S_{t,\mathrm{test}}^{(m)}$
denote the resulting task scores for method $m$.
We average over all 30 tasks and report scores on a $0$--$100$ scale:
\[
    \operatorname{DevScore}^{(m)}
    =
    \frac{100}{|\mathcal{T}|}
    \sum_{t \in \mathcal{T}} S_{t,\mathrm{dev}}^{(m)},
    \qquad
    \operatorname{TestScore}^{(m)}
    =
    \frac{100}{|\mathcal{T}|}
    \sum_{t \in \mathcal{T}} S_{t,\mathrm{test}}^{(m)}.
\]

\paragraph{Submission validity.}
Let $v_t^{(m)}$ indicate whether the final submission of method $m$
on task $t$ successfully completes test evaluation and satisfies
the benchmark's validity requirements.
The valid-submission rate is
\[
    \operatorname{ValidSub}^{(m)}
    =
    \frac{1}{|\mathcal{T}|}
    \sum_{t \in \mathcal{T}} v_t^{(m)}.
\]

\paragraph{Comparison with human baselines.}
All comparisons with human baselines use test results.
For task $t$, let $\mathcal{B}_t$ denote the set of provided human
baselines, each evaluated on the same test split and scored using
the same procedure as the agent submission.
We define the comparison thresholds as
\[
    \tau_t^{\mathrm{Any}}
    = \min_{b \in \mathcal{B}_t} S_{t,\mathrm{test}}^{(b)},
    \qquad
    \tau_t^{\mathrm{Median}}
    = \operatorname{median}_{b \in \mathcal{B}_t}
      S_{t,\mathrm{test}}^{(b)},
    \qquad
    \tau_t^{\mathrm{All}}
    = \max_{b \in \mathcal{B}_t} S_{t,\mathrm{test}}^{(b)}.
\]
\textbf{Beat Any}, \textbf{Beat Median}, and \textbf{Beat All}
report the percentages of tasks on which a valid final submission
strictly exceeds the corresponding threshold:
\[
    \operatorname{Beat}_{c}^{(m)}
    =
    \frac{1}{|\mathcal{T}|}
    \sum_{t \in \mathcal{T}}
    v_t^{(m)}
    \mathbf{1}\!\left[
        S_{t,\mathrm{test}}^{(m)} > \tau_t^{c}
    \right],
    \qquad
    c \in \{\mathrm{Any}, \mathrm{Median}, \mathrm{All}\}.
\]
Here, $\mathbf{1}[\cdot]$ is the indicator function.
All rates are reported as percentages, with all 30 tasks included
in the denominator.
Invalid submissions and ties do not count as wins.

\subsection{Evaluation on the Three Research Tasks}
\label{app:three_task_metrics}

We first define the task metrics independently of the evaluation setting.

\paragraph{Validation bits per byte.}
Let $y_i$ be the $i$-th validation token, let
\[
    \ell_i
    =
    -\log p_\theta(y_i\mid y_{<i})
\]
be its negative log-likelihood in nats, and let $b_i$ be the number of bytes
represented by that token. Tokens with zero byte length, such as special
control tokens, are excluded. Validation BPB is
\[
    \operatorname{BPB}
    =
    \frac{\sum_{i:b_i>0}\ell_i}
         {\log 2\sum_{i:b_i>0} b_i}.
\]
It is the average number of bits required to encode one byte of validation
text under the trained model, so lower values indicate better language
modeling.

\paragraph{Kernel speedup.}
Let $\bar{t}_{\mathrm{ref}}$ and $\bar{t}_{\mathrm{cand}}$ denote the mean
CUDA-event runtimes of the eager reference and candidate implementation,
respectively. The KernelBench metric is
\[
    \operatorname{Speedup}
    =
    \begin{cases}
    \displaystyle
    \frac{\bar{t}_{\mathrm{ref}}}
         {\bar{t}_{\mathrm{cand}}},
         & \text{if all correctness checks pass},\\[8pt]
    0,   & \text{otherwise}.
    \end{cases}
\]
A value greater than one indicates that the candidate is faster than the
reference. Failed compilation, runtime errors, shape mismatches, or numerical
correctness failures receive zero.

\paragraph{Macro Pearson correlation.}
For trait $j$, let $y_{j,n}$ and $\widehat{y}_{j,n}$ be the ground-truth and
predicted phenotypes of evaluation observation $n$. Its Pearson correlation is
\[
    r_j
    =
    \frac{
        \sum_{n=1}^{N_j}
        (y_{j,n}-\bar{y}_j)
        (\widehat{y}_{j,n}-\bar{\widehat{y}}_j)
    }{
        \sqrt{
        \sum_{n=1}^{N_j}(y_{j,n}-\bar{y}_j)^2
        }
        \sqrt{
        \sum_{n=1}^{N_j}
        (\widehat{y}_{j,n}-\bar{\widehat{y}}_j)^2
        }
    }.
\]
The G2P score is the unweighted mean over the eight traits:
\[
    \operatorname{MacroPearson}
    =
    \frac{1}{8}\sum_{j=1}^{8}r_j.
\]
Higher values indicate that the predictions better preserve phenotype
variation across hybrid--environment observations.

\paragraph{Sample-efficiency speedup.}
We quantify search efficiency by comparing the number of experiments required
by Janus and a competing method to reach the same performance threshold. Let
$y_{m,t}(k)$ denote the score obtained by method $m$ on task $t$ after
experiment $k$, where $k=0$ corresponds to the initial solution. To accommodate
metrics with different optimization directions, we define
\[
    d_t =
    \begin{cases}
        -1, & \text{if lower scores are better},\\
        +1, & \text{if higher scores are better}.
    \end{cases}
\]
The direction-aligned best-so-far trajectory is
\[
    F_{m,t}(k)
    =
    \max_{0\leq j\leq k} d_t y_{m,t}(j).
\]
An unsuccessful experiment consumes one experiment but does not update
$F_{m,t}$. For a direction-aligned performance threshold $u$, the first-hitting
time of method $m$ is
\[
    \tau_{m,t}(u)
    =
    \min\left\{k:F_{m,t}(k)\geq u\right\}.
\]

For Janus $J$ and baseline $b$, we evaluate thresholds over their jointly
reachable interval
\[
    \mathcal{U}_{t,b}
    =
    \left(u_{t,b}^{\mathrm{lo}},u_{t,b}^{\mathrm{hi}}\right],
\]
where
\[
    u_{t,b}^{\mathrm{lo}}
    =
    \max\{F_{J,t}(0),F_{b,t}(0)\},
    \qquad
    u_{t,b}^{\mathrm{hi}}
    =
    \min\{F_{J,t}(N_t),F_{b,t}(N_t)\}.
\]
The lower endpoint is excluded because both methods may have zero hitting time
at the initial score. The threshold-wise sample-efficiency speedup is
\[
    \rho_{t,b}(u)
    =
    \frac{\tau_{b,t}(u)}{\tau_{J,t}(u)}.
\]
Thus, $\rho_{t,b}(u)>1$ indicates that the baseline requires more experiments
than Janus to reach threshold $u$.
The \emph{average speedup} reported under \emph{Avg.} in Table~2 is
\[
    \operatorname{Avg}_{t,b}
    =
    \frac{1}{
        u_{t,b}^{\mathrm{hi}}-u_{t,b}^{\mathrm{lo}}
    }
    \int_{u_{t,b}^{\mathrm{lo}}}^{u_{t,b}^{\mathrm{hi}}}
    \frac{\tau_{b,t}(u)}{\tau_{J,t}(u)}
    \,\mathrm{d}u.
\]
This quantity averages the experiment-count ratio uniformly over all jointly
reachable performance thresholds, rather than over experiment indices. The
\emph{maximum speedup} reported under \emph{Max.} is
\[
    \operatorname{Max}_{t,b}
    =
    \operatorname*{ess\,sup}_{u\in\mathcal{U}_{t,b}}
    \frac{\tau_{b,t}(u)}{\tau_{J,t}(u)}.
\]
The essential supremum excludes ratios that occur only at an isolated,
zero-width threshold.


\section{Implementation Details}
\label{app:implementation}

\paragraph{Common setup.}
Each experimental run is assigned a single NVIDIA A100 PCIe GPU with 40\,GB of
memory. The primary comparisons use DeepSeek-V4-Flash as the common language
model backbone; experiments with the other backbones reported in the main text
serve as supplementary robustness comparisons. Within each common-backbone
comparison, we keep the model endpoint, reasoning configuration, task prompt,
initial workspace, action permissions, evaluator, and research budget fixed.
Different methods are not allowed to share trajectories, intermediate
candidates, evaluator outputs, or memories.

All evaluations are performed by immutable external harnesses. A candidate is
accepted only if it modifies files inside the declared editable surface and
completes without violating the task-specific resource or correctness
constraints. Compilation failures, timeouts, invalid outputs, and non-finite
metrics are recorded as failed experiments rather than silently discarded.

\paragraph{MLS-Bench-Lite.}
Each agent is given a 5-hour wall-clock research budget per task.
Since our experiments use A100 GPUs, we relax the default
per-task evaluator timeouts to accommodate hardware-dependent
differences in execution speed relative to the H100 setup
reported in MLS-Bench~\cite{lyu2026mls}.
This adjustment allows the prescribed evaluation workloads
to complete without premature termination caused by slower
execution on our hardware.
The same timeout policy is applied to all compared methods,
while the agent's research budget remains unchanged. Under the fixed wall-clock budget, the world model can select top-ranked proposals for parallel execution on the allocated
GPU when computational resources permit, improving the utilization of the available research time.
To further ensure comparability across methods and backbones, we predefine and hold fixed a common validation split for each task, rather than allowing individual agents to construct their own splits, so that differences in validation performance more directly reflect their ability to optimize under the same feedback.

We retain the benchmark's task-specific validity constraints.
For tasks involving editable model components, the original
parameter-budget checks remain enforced, and submissions
exceeding the prescribed capacity ceiling are rejected.
Edits remain restricted to the designated component, and
shared training-protocol settings, such as training epochs
and batch size, remain within their protected ranges.
Method-defining hyperparameters are editable only where
permitted by the task.
The evaluation data, settings, seed policies, correctness
checks, and scoring procedures are also preserved.
The additional evaluator runtime therefore accommodates
the original workload while maintaining the benchmark's
controls on model capacity and permissible modifications.

\paragraph{nanoGPT Training.}
Our experiments use a fixed training budget of 1,200 seconds
per candidate, excluding startup, compilation, and the initial
warm-up iterations.
The sequence length is 2,048, and the tokenizer vocabulary
contains 8,192 entries.
The validation set is the pinned final shard of the
ClimbMix-400B-Shuffle corpus, with BPB evaluated over
20,971,520 validation tokens. Training and validation use the same fixed tokenizer,
dataloader, document-packing procedure, and validation shard for all methods.
Mixed-precision training uses BF16 autocasting,
while metric accumulation is performed by the protected evaluation routine.
Only \texttt{train.py} is editable; the data preparation and
\texttt{evaluate\_bpb} implementations are read-only.

\paragraph{MinGPT Causal Attention Optimization.}
For the MinGPT Causal Attention optimization task from
KernelBench Level~3, Task~43, the official workload uses
a batch size $B$ of 128, a sequence length $T$ of 512,
and an embedding dimension $C$ of 768.
The model has eight attention heads, supports a maximum
sequence length of 1,024, and uses zero attention and
residual dropout.
Inputs and outputs are FP32 tensors.
Parameters are initialized deterministically, and correctness
is checked on five freshly generated inputs using fixed seeds.
On every trial, the candidate output must match the reference
tensor shape and pass an elementwise closeness check with
absolute and relative tolerances both set to 0.0001.

Performance is measured with CUDA events after three warm-up executions.
We clear the allocator cache and perturb the L2-cache state before timed
executions, discard the first timing observation, and report the mean of
100 retained trials. The clean eager-reference latency is frozen at
$32.6$\,ms and is used as the numerator of the speedup score, preventing a
candidate from altering its own reference timing through global backend flags.
Only \texttt{solution.py} may be modified, and it must expose a top-level
\texttt{ModelNew} class compatible with the reference constructor.

\paragraph{Genotype-to-Phenotype.}
The genotype-to-phenotype (G2P) prediction evaluator exposes a phenotype-independent genotype matrix containing
7,577 single-nucleotide polymorphism (SNP) dosages per hybrid and two fixed representations of each environment:
a basic statistical summary and a richer temporal summary of the daily weather
sequence. The latter includes global statistics, quantiles, temporal trends,
and stage-wise summaries. The split is shared across all eight traits so that
multi-trait models cannot obtain information through inconsistent
trait-specific partitions.

For the default strict-M3 protocol, hybrids and environments are partitioned
into disjoint training, validation, and test groups with a target observation
ratio of $70\%/10\%/20\%$. The resulting entity partitions contain
$2{,}232/844/1{,}193$ hybrids and $97/37/51$ environments, respectively.
The evaluator withholds phenotype targets for all evaluation rows.
Only \texttt{train.py} is editable, and all preprocessing influenced by labels
or sample distributions must be fitted from the supplied training pool. Each
candidate evaluation has a 900-second timeout. The principal score is macro
Pearson; per-trait Pearson, Spearman correlation, and within-environment
Pearson correlation are retained only as diagnostic statistics.

\section{Out-of-Distribution Evaluation Setup and Analysis}
\label{app:ood}


We define one controlled distribution shift for each of the three research
tasks. The OOD settings are specified before evaluating the final candidates. Each shifted setting retains the original evaluator and primary
metric while changing one scientifically meaningful axis.

\paragraph{Kernel workload shift.}
The default attention workload uses a batch size $B$ of 128
and a sequence length $T$ of 512.
In the shifted setting, we reduce the batch size to 64
and increase the sequence length to 1,024.
The embedding dimension $C$ remains 768, with eight attention
heads, FP32 precision, and unchanged attention semantics.
Both settings contain 65,536 input tokens per batch, but the
longer sequences increase the quadratic attention workload
and alter the balance among projection, attention, memory,
and kernel-launch costs.
Correctness is checked on five inputs with absolute and
relative tolerances both set to 0.0001.
We then perform three warm-up runs followed by 100 CUDA-event
timing trials.
The eager reference latency in this setting is 45.9\,ms.

\paragraph{Cross-corpus transfer.}
For nanoGPT, we replace the original ClimbMix corpus with
TinyStories~\cite{eldan2023tinystories}.
We construct fixed, document-disjoint training and evaluation
partitions for the new corpus.
Each discovered method is frozen and retrained from scratch
on the new training partition, without additional method search
or hyperparameter tuning.
The 1,200-second training budget, original tokenizer,
2,048-token context length, hardware, random-seed policy,
and BPB metric remain unchanged.
We evaluate the initial and final methods, together with the
unmodified baseline, under this shared setting to assess
whether improvements found on the original corpus transfer
to a different text distribution.

\paragraph{Cross-variety transfer.}
For G2P, we evaluate method transfer to a separate cohort of
maize hybrids excluded from the original research process.
We retain each discovered method's modeling design and
hyperparameters, retrain its parameters on the new cohort's
training partition, and evaluate it on a test partition
containing disjoint hybrid identities.
The SNP and daily-weather feature definitions, eight target
traits, environmental evaluation protocol, training budget,
and macro-Pearson metric remain unchanged.
The initial and final methods and the unmodified baseline
are evaluated on the same partitions.
This setting examines whether the discovered modeling
improvements remain effective across different hybrid
cohorts without further research or tuning on the target cohort.

\paragraph{Cross-task score calibration.}
Because the native evaluation functionals of the three tasks differ in scale,
orientation, and physical interpretation, we first transform each raw
measurement into a task-oriented gain and then calibrate that gain against a
fixed reference scale. Let
\[
\Pi_{\mathbb{R}_{\geq 0}}(x)
\coloneqq
\underset{z\in\mathbb{R}_{\geq 0}}{\operatorname{argmin}}
\,|z-x|
=
\max\{0,x\}
\]
denote projection onto the non-negative real line. For a task
\(\tau\in\mathcal{T}
=\{\mathrm{kernel},\mathrm{nanoGPT},\mathrm{G2P}\}\),
we define the normalized score functional
\[
\mathcal{Q}_{\tau}
\bigl(\mathsf{M}_{\tau};\mathsf{M}^{(0)}_{\tau}\bigr)
\coloneqq
100\,
\frac{
\Pi_{\mathbb{R}_{\geq 0}}
\left(
\mathcal{G}_{\tau}
\bigl(\mathsf{M}_{\tau};\mathsf{M}^{(0)}_{\tau}\bigr)
\right)
}{
\Lambda_{\tau}
},
\]
where \(\mathsf{M}_{\tau}\) is the candidate's raw metric,
\(\mathsf{M}^{(0)}_{\tau}\) is the corresponding setting-specific reference
metric, \(\mathcal{G}_{\tau}\) is a direction-aware gain transformation, and
\(\Lambda_{\tau}>0\) is a task-specific calibration constant.

The gain transformations are instantiated as
\[
\mathcal{G}_{\tau}
\bigl(\mathsf{M}_{\tau};\mathsf{M}^{(0)}_{\tau}\bigr)
=
\begin{cases}
\dfrac{t_{0}}{t},
&
\tau=\mathrm{kernel},
\\[1.1em]
b_{0}-b,
&
\tau=\mathrm{nanoGPT},
\\[0.6em]
r-r_{0},
&
\tau=\mathrm{G2P},
\end{cases}
\]
where \(t_{0}/t\) is the candidate's speedup relative to the
setting-specific reference runtime, \(b_{0}-b\) is its absolute reduction in validation BPB,, and \(r-r_{0}\) is its absolute improvement in macro
Pearson correlation.

The calibration constants are fixed once and shared across all methods,
optimization stages, and evaluation distributions. Consequently, a four-fold KernelBench speedup, an absolute nanoGPT BPB
reduction of \(0.05\), and an absolute G2P macro-Pearson improvement of
\(0.15\) each define the task-specific 100-point reference level. The
projection suppresses negative directional gains, while no upper saturation
is applied; hence scores greater than 100 remain admissible whenever a method
surpasses the corresponding calibration target.

Finally, the cross-task aggregate is defined as the expectation under the
uniform measure over the task index set:
\[
\mathcal{Q}_{\mathrm{aggregate}}
\coloneqq
\mathbb{E}_{\tau\sim\operatorname{Unif}(\mathcal{T})}
\left[
\mathcal{Q}_{\tau}
\right]
=
\frac{1}{|\mathcal{T}|}
\sum_{\tau\in\mathcal{T}}
\mathcal{Q}_{\tau}.
\]

This calibration places the heterogeneous task metrics on commensurate
numerical scales, thereby making their uniformly weighted arithmetic average
meaningful despite differences in their native units and optimization
directions.

\paragraph{OOD generalization analysis.}
Figure~\ref{fig:framework_analysis}(b) compares the aggregate ID and OOD
scores of the initial and final research solutions produced by
each method. The OOD settings are defined independently of the
research process, and the discovered methods are evaluated under
the corresponding shifts without further research or tuning.
Therefore, movement from the initial to the final point reflects
whether improvements obtained during research remain effective
beyond the original optimization setting.

JANUS exhibits the strongest transfer among the compared methods.
Its final solution reaches the highest aggregate score under both
ID and OOD evaluation, showing that the improvements discovered
during research are not confined to the distribution on which
they were developed. Importantly, its final point remains close
to the diagonal, indicating that the gain achieved on the original
tasks is largely preserved after distribution shift. In contrast,
several baselines show a larger separation between their final
ID and OOD scores, suggesting that part of their research-time
improvement is more sensitive to changes in workload, corpus, or
evaluation cohort.

The same behavior is particularly meaningful because the three
OOD settings exercise different forms of transfer: a workload
shift for the attention kernel, cross-corpus transfer for nanoGPT,
and cross-variety transfer for G2P. Thus, JANUS's robustness does
not depend on a single type of distribution shift. Together, these
results indicate that JANUS tends to discover research solutions
whose improvements persist under changes to the evaluation
distribution, rather than narrowly optimizing for the original
research setting.

\section{Analysis of Baseline Imitation}
\label{app:baseline_imitation}

\paragraph{Similarity-weighted baseline performance.}
We further analyze whether the performance obtained by each
research agent can be explained by reproducing solutions from
strong human-designed baselines. For each evaluated submission,
we measure its code similarity to every baseline $i$ using
1/3-gram Jaccard similarity, denoted by $s_i$. Let $\beta_i$
denote the performance of baseline $i$. We define the
similarity-weighted baseline performance as
\begin{equation}
    q =
    \frac{\sum_i s_i \beta_i}
         {\sum_i s_i}.
\end{equation}
Intuitively, $q$ is high when a submission more closely resembles
high-performing baselines, and low when its similarity is
concentrated on weaker baselines or is less aligned with the
strongest known solutions. To make the quantities comparable
across tasks, both $q$ and the corresponding agent performance
are normalized by the best baseline performance of that task.

For each research agent, we fit the relationship between its
normalized performance and similarity-weighted baseline
performance across evaluated submissions, as shown in
Figure~\ref{fig:reward_similarity}. A strong positive dependence
would indicate that high-performing submissions tend to coincide
with greater resemblance to stronger baselines. In contrast,
JANUS achieves comparatively high fitted performance even when
$q$ is low and exhibits a shallower fitted slope than the other
agents. Its performance therefore varies less with the quality
of the baselines that its solutions resemble.

The weaker dependence on similarity-weighted baseline quality
highlights JANUS's ability to improve beyond the strongest
existing implementations. Rather than relying on close resemblance
to high-performing baselines, JANUS can identify effective
alternative solutions throughout the research process, enabling
broader exploration and a higher potential performance ceiling.

\section{Analysis of Adaptive Research Strategies}
\label{app:strategy_analysis}

\paragraph{Trajectory and assessment.}
We analyze all selected candidates with completed
formal evaluations in the nanoGPT trajectory, ordered by
their actual execution sequence.
For each candidate $t$, we recover its proposal text $h_t$
as recorded when it was proposed.
We use GPT-5.6 Sol as an LLM judge to assess the proposal's
research emphasis under a shared rubric.
The assessment excludes the candidate's own evaluation
outcome and all subsequent results.
References to earlier experiments already present in the
proposal are retained, as they describe the evidence
available when the proposal was formulated.
The analysis therefore characterizes the stated intent of
selected proposals independently of their eventual success.

\paragraph{Strategy dimensions.}
We characterize each proposal along three non-exclusive
dimensions.
\textbf{Exploitation} measures how strongly it builds on
previously supported mechanisms to pursue immediate
improvement, such as refining a promising configuration
or combining effective modifications.
\textbf{Information gain} measures its emphasis on reducing
uncertainty about which proposal or research direction
should be pursued next, through controlled comparisons,
replication, or mechanism isolation.
\textbf{Exploration} measures its emphasis on investigating
unverified mechanisms or alternative approaches that may
open promising directions for subsequent research.
The resulting scores are denoted by $X_t,I_t,E_t\in[0,1]$.
They need not sum to one, since a proposal can pursue several
objectives simultaneously.
For example, testing a new mechanism can both reduce
uncertainty about its effectiveness and reveal promising
follow-up directions.

\paragraph{Temporal smoothing.}
To reveal changes in research emphasis while reducing
experiment-to-experiment fluctuations, we smooth each strategy
dimension using a centered seven-experiment moving average.
Let $s_{t,d}$ denote the original score at experiment $t$
for dimension $d\in\{X,I,E\}$.
The smoothed score is
\[
    \overline{s}_{t,d}
    =
    \frac{1}{7}
    \sum_{k=-3}^{3}s_{\pi(t+k),d},
    \qquad
    \pi(j)=\min\{N,\max\{1,j\}\},
\]
where $N=50$.
The index mapping repeats endpoint values near the trajectory
boundaries.
In Figure~\ref{fig:framework_analysis}(c), faint colored lines
and points show the original per-proposal scores, while thick
curves show their moving averages.

\paragraph{Normalized performance trace.}
The purple staircase provides experimental context by showing
the cumulative improvement in validation BPB.
Let $b_0$ denote the initial baseline BPB and $b_t$ the
evaluation result of candidate $t$.
Since lower BPB is better, the incumbent performance is
\[
    b_t^\star=\min\{b_0,b_1,\ldots,b_t\}.
\]
We normalize its improvement relative to the total improvement
achieved within the displayed trajectory:
\[
    Z_t
    =
    \frac{b_0-b_t^\star}{b_0-b_N^\star}.
\]
This maps the initial baseline to zero and the final incumbent
to one.
The staircase retains the current value until the next
incumbent improvement.


\end{document}